\documentclass{article} % For LaTeX2e
\usepackage{iclr2027_conference,times}

\usepackage{amsmath,amsfonts,bm}

\def\eqref#1{equation~\ref{#1}}
\def\1{\bm{1}}

\DeclareMathAlphabet{\mathsfit}{\encodingdefault}{\sfdefault}{m}{sl}
\SetMathAlphabet{\mathsfit}{bold}{\encodingdefault}{\sfdefault}{bx}{n}

\usepackage{hyperref}
\usepackage{url}
\usepackage{graphicx}
\usepackage{booktabs}
\usepackage{multirow}
\usepackage{xcolor}
\usepackage{subcaption}
\usepackage{wrapfig}
\usepackage{enumitem}
\usepackage{makecell}
\usepackage{bm}

\title{Rethinking Universal Time Series Foundation Models from Unified Representations to Shared Dynamics (UniTFU)}

\author{
Shuxin Li\textsuperscript{1},
Yingjie Zhou\textsuperscript{1}\thanks{ Corresponding author.},
Le Zhang\textsuperscript{2},
Jingyuan Xia\textsuperscript{3},
Shengxi Li\textsuperscript{4},
Ce Zhu\textsuperscript{5},
Anderson Rocha\textsuperscript{6} \\
\textsuperscript{1}Sichuan University, Chengdu, China;\\
\textsuperscript{2}Hohai University, Nanjing, China;\\
\textsuperscript{3}National University of Defense Technology, Changsha, China;\\
\textsuperscript{4}Beihang University, Beijing, China;\\
\textsuperscript{5}University of Electronic Science and Technology of China, Chengdu, China;\\
\textsuperscript{6}Universidade Estadual de Campinas, Campinas, Brazil;\\
\texttt{lishuxin@stu.scu.edu.cn, yjzhou09@gmail.com, zhangleuestc@gmail.com, }\\
\texttt{j.xia10@nudt.edu.cn, LiShengxi@buaa.edu.cn, eczhu@uestc.edu.cn,}\\
\texttt{anderson.rocha@gmail.com}\\
}

\iclrfinalcopy

\begin{document}

\maketitle
\lhead{}

\begin{abstract}
Time series foundation models (TSFMs) have attracted increasing attention in cross-domain forecasting. Existing works mainly rely on large-scale pre-training and general modeling paradigms. However, the former heavily relies on the diversity and domain coverage of pre-training data. Meanwhile, the latter learns unified representations across heterogeneous domains without explicitly identifying the temporal regularities shared among them, limiting robust cross-domain generalization. This work explores cross-domain generalization by characterizing commonalities in recurring local dynamics across domains, rather than relying solely on implicitly learned temporal representations. To capture such cross-domain commonalities, we propose \textbf{Time Series Foundation Units (TFUs)} as reusable building blocks discovered from multi-scale temporal segments, each characterizing a recurring local dynamic pattern shared across domains. Building on TFUs, we develop a unified time series foundation model (\textbf{UniTFU}) that explicitly reuses and composes shared temporal regularities across domains for forecasting. Experiments on seven public datasets show that UniTFU substantially outperforms existing TSFMs in cross-domain forecasting and can efficiently adapt to target-domain dynamics by updating its reusable components with limited target data. UniTFU also achieves strong forecasting performance compared with state-of-the-art methods.

\end{abstract}

\section{Introduction}
Time series forecasting plays a crucial role across many domains, including weather forecasting \citep{NEURIPS2021_bcc0d400}, electricity monitoring \citep{7998331}, and energy management \citep{6714975}. As forecasting applications expand across diverse domains, specialized models designed for individual domains \citep{PatchTST,iTransformer,yao2025towards} struggle to scale across domains. Inspired by the success of foundation models (FMs) in natural language processing (NLP) \citep{zhao2023survey} and computer vision (CV) \citep{10834497}, time series foundation models (TSFMs) have emerged as a promising paradigm for cross-domain forecasting. By learning general temporal representations from large-scale time series data, TSFMs enable a single model to perform forecasting across diverse domains \citep{das2023decoder,liu2024timer,wu2025aurora}.

A major trend in current TSFM research is to build general forecasters with unified modeling capability, enabling consistent cross-domain forecasting performance \citep{aksu2024gift}. Existing TSFMs generally pursue cross-domain generalization from two perspectives: 1) large-scale pre-training and 2) general modeling paradigms. TSFMs based on large-scale pre-training \citep{Sundial,jin2024time} learn shared temporal patterns from large-scale, multi-domain time series corpora, enabling cross-domain generalization. However, their generalization largely depends on the diversity and domain coverage of pre-training data \citep{zhang2025does}, limiting robust performance on highly heterogeneous or underrepresented domains. TSFMs with general modeling paradigms \citep{liu2025timer,chenggtm} accommodate different temporal scales, variable dimensions, and forecasting tasks within a unified representation space \citep{woo2024unified}, but such unified representations do not explicitly identify the temporal regularities shared across domains \citep{otovic2022intra,ning2026ts}, thereby limiting robust cross-domain generalization. Overall, existing approaches largely obtain cross-domain generalization through data-driven representation learning or unified modeling, while recurring local dynamics shared across domains remain insufficiently characterized and explicitly reused. This raises a key question: \textit{How can TSFMs characterize the commonalities in temporal evolution across domains for robust cross-domain forecasting?}

\begin{figure}[t]
    \centering
    \includegraphics[width=\linewidth]{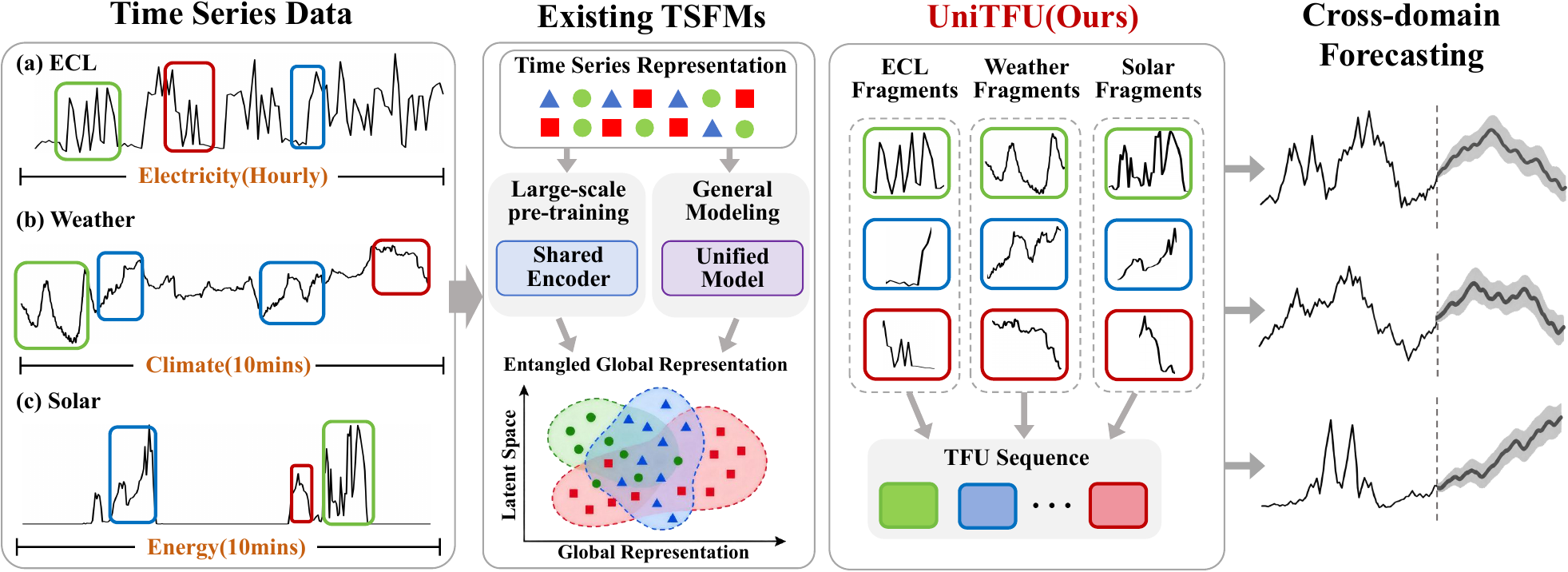}
    \caption{Conceptual comparison of existing TSFMs and UniTFU. Existing TSFMs mainly rely on implicit temporal representations, whereas UniTFU explicitly organizes recurring local dynamics across domains into shared TFUs for reuse and composition in cross-domain forecasting.}
    \label{fig:introduction}
    \vspace{-3pt}
\end{figure}

In this work, we introduce a new modeling perspective that grounds TSFMs' general forecasting capability in discovering, reusing, and composing shared local dynamic structures across domains. Accordingly, we propose the \textbf{Time Series Foundation Unit (TFU)} as a reusable and composable prototype that characterizes recurring local dynamic patterns within finite temporal segments. As illustrated in Figure~\ref{fig:introduction}, existing TSFMs primarily rely on implicit temporal representations, whereas TFUs explicitly organize recurring local dynamics into reusable units for cross-domain forecasting. Such units may capture recurring local behaviors such as rising, falling, oscillatory, or transition patterns. Rather than being manually predefined, TFUs are automatically discovered from large-scale time series data and continuously merged and updated into a shared dictionary. Complex temporal dynamics can then be represented as compositions of these shared units, enabling explicit reuse of temporal knowledge across domains.

Building on TFUs, we develop \textbf{UniTFU}, a unified time series foundation model that leverages shared TFUs to guide temporal dependency modeling and refine forecasting across domains. Specifically, UniTFU incorporates unit-level TFUs and their aggregated scene context into the attention mechanism to guide temporal dynamics modeling. Experimental results show that UniTFU exhibits robust cross-domain generalization in forecasting and effectively adapts to target-domain dynamics with limited data. UniTFU also achieves strong performance on standard forecasting benchmarks. Our contributions are summarized as follows:

% Experimental results show robust cross-domain generalization and further show that TFUs learned from source domains can be effectively reused and adapted to target domains with limited target data. 

% Experimental results show that UniTFU exhibits robust cross-domain generalization in forecasting and effectively adapts to target-domain dynamics with limited data.

% UniTFU also achieves strong performance on standard forecasting benchmarks. Our contributions are summarized as follows:

\begin{itemize}[leftmargin=1.5em]
    \item{We introduce TFUs as reusable and composable prototypes of recurring local dynamics discovered from multi-scale temporal segments. TFUs provide a unit-level perspective on shared temporal regularities across domains, offering a foundation for Universal time series forecasting.} 
    \item{We develop UniTFU, a TSFM that explicitly integrates the discovery, reuse, and composition of TFUs into forecasting, enabling shared local dynamics to be leveraged across domains.} 
    \item{We systematically evaluate UniTFU in cross-domain settings and on standard forecasting benchmarks, demonstrating robust cross-domain generalization and strong forecasting performance. Further analysis shows that TFUs learned from source domains can be reused in target domains and further adapted with limited target data.}
\end{itemize}

\section{Related Work}

\subsection{Time Series Foundation Models for Forecasting}
Existing TSFMs mainly follow two complementary routes: large-scale pre-training and general modeling paradigms. Large-scale pretrained TSFMs learn general temporal representations from multi-domain corpora for zero-shot forecasting and cross-domain adaptation. Representative methods include TimesFM~\citep{das2023decoder}, Time-MoE~\citep{shi2025time}, MOIRAI~\citep{woo2024unified}, Chronos~\citep{ansari2024chronos}, and Sundial~\citep{Sundial}. Recent works such as Chatime~\citep{wang2025chattime} and Aurora~\citep{wu2025aurora} further incorporate multimodal information into TSFM pre-training for heterogeneous domains. However, their generalization largely depends on pre-training diversity and domain coverage, limiting robustness on underrepresented target domains.

Another line of work develops general modeling paradigms for general-purpose time series forecasting. Early deep forecasting models, such as DLinear~\citep{zeng2023transformers}, PatchTST~\citep{PatchTST}, iTransformer~\citep{iTransformer}, and PhaseFormer~\citep{niu2026phaseformer}, introduce general modeling structures for long-term and multivariate forecasting. These methods promote unified forecasting beyond task-specific architectures. Recent TSFMs further incorporate unified architectures into foundation-model design, including Timer-XL~\citep{liu2025timer}, Timer~\citep{liu2024timer}, and GTM~\citep{chenggtm}. Although these methods improve adaptability across diverse forecasting scenarios, their cross-domain generalization mainly relies on unified architectures, representations, or prediction procedures, which do not explicitly identify and reuse the recurring local dynamic structures shared across domains, making robust cross-domain generalization challenging.

\begin{table}[h]
\centering
\caption{Comparison with representative methods for structured representation learning.}
\label{tab:tfu_positioning}
{\footnotesize\setlength{\tabcolsep}{3pt}
\resizebox{\linewidth}{!}{%
\begin{tabular}{@{}lcccccccc@{}}
\toprule
Mechanism & ShapeX & L-MAP & TOTEM & ProtoTS & E2E-SAE & WQ4TS & APT
          & \textbf{UniTFU} \\
\midrule
Subsequence matching
  & $\checkmark$ & $\checkmark$ & $\checkmark$ & $\times$
  & $\times$ & $\times$ & $\times$ & $\checkmark$ \\
Learned dictionary
  & $\checkmark$ & $\times$ & $\checkmark$ & $\checkmark$
  & $\checkmark$ & $\times$ & $\checkmark$ & $\checkmark$ \\
Cross-domain forecasting
  & $\times$ & $\times$ & $\checkmark$ & $\times$
  & $\times$ & $\checkmark$ & $\times$ & $\checkmark$ \\
Scene-conditioned attention
  & $\times$ & $\times$ & $\times$ & $\times$
  & $\times$ & $\times$ & $\times$ & $\checkmark$ \\
Prototype birth \& consolidation
  & $\times$ & $\times$ & $\times$ & $\times$
  & $\times$ & $\times$ & $\times$ & $\checkmark$ \\
\bottomrule
\end{tabular}}}
\end{table}

\subsection{Local Temporal Representation Learning for Time Series}
% Local temporal patterns have long been studied as compact representations of time series. Shapelet-based methods identify discriminative subsequences for classification \citep{ye2009time,grabocka2014learning}, while motif discovery searches for recurring subsequences through similarity matching \citep{mueen2009exact}. Beyond direct subsequence discovery, dictionary learning represents signals through sparse combinations of learned atoms \citep{mairal2009online}, whereas symbolic approaches encode local subsequences into reusable symbolic words for classification \citep{schafer2015boss}. Neural approaches further introduce learned codebooks and prototype representations. 
% VQ-based models quantize continuous latent embeddings using a learned discrete codebook \citep{van2017neural}, while prototypical networks represent each class by the mean embedding of its support examples \citep{snell2017prototypical}. TOTEM \citep{talukder2024totem} learns a VQ-based tokenizer independently of downstream models and reuses the resulting tokenization across domains and tasks.

% In this work, we introduce TFUs as reusable prototypes of recurring local dynamics and develop UniTFU. Rather than learning a discrete tokenization through a fixed-size codebook, UniTFU explicitly discovers and organizes recurring local dynamics into reusable TFUs, which are adaptively updated, expanded, and consolidated during forecasting training and composed across multiple scales to guide cross-domain forecasting.

Local temporal patterns have long been studied as compact representations of time series. Shapelet-based methods identify discriminative subsequences for classification \citep{ye2009time,grabocka2014learning}, while ShapeX \citep{NEURIPS2025_26300d27} uses learned shapelets for classification explanations. Motif discovery identifies recurring subsequences \citep{mueen2009exact}, and L-MAP \citep{Lin_Wang_Hong_Li_2026} learns multivariate motif similarities. Dictionary learning represents signals through sparse combinations of learned atoms \citep{mairal2009online}, while E2E-SAE \citep{NEURIPS2024_c212c1b8} learns functionally relevant dictionaries of neural activations. Symbolic methods encode subsequences as words \citep{schafer2015boss}, and WQ4TS \citep{10.1145/3788674} explores cross-domain transfer through a shared wavelet-based spectral representation. Neural approaches further employ VQ codebooks \citep{van2017neural} and prototype representations \citep{snell2017prototypical}. TOTEM \citep{talukder2024totem} learns a VQ-based tokenizer independently of downstream models and reuses the resulting tokenization across domains and tasks. ProtoTS \citep{peng2026protots} learns hierarchical prototypes for interpretable forecasting, while APT \citep{Li_Shao_Yu_Fu_Sun_Xu_Wang_2026} uses timestamp-conditioned prototypes for affine adaptation.

In this work, we introduce TFUs as reusable prototypes of recurring local dynamics and develop UniTFU. While TFU assignment follows a VQ-like nearest-prototype principle, UniTFU composes multi-scale matches into temporally aligned representations, integrates local TFUs and scene context into forecasting attention, and adaptively refines, expands, and consolidates the shared dictionary. Table~\ref{tab:tfu_positioning} summarizes these distinctions.

% In this work, we introduce TFUs as reusable prototypes of recurring local dynamics and develop UniTFU, which maintains a shared adaptive dictionary and composes matched prototypes into a temporally aligned sequence for cross-domain forecasting.

% In this work, we introduce TFUs as reusable prototypes of recurring local dynamics and develop UniTFU. Unlike learning a separate tokenizer with a fixed-size codebook, UniTFU adaptively updates, adds, and consolidates prototypes during forecasting training, and composes matched multi-scale prototypes at their covered token positions to guide cross-domain forecasting.

\section{Model}
\label{sec:model}

\begin{figure}[t]
    \centering
    \includegraphics[width=\linewidth]{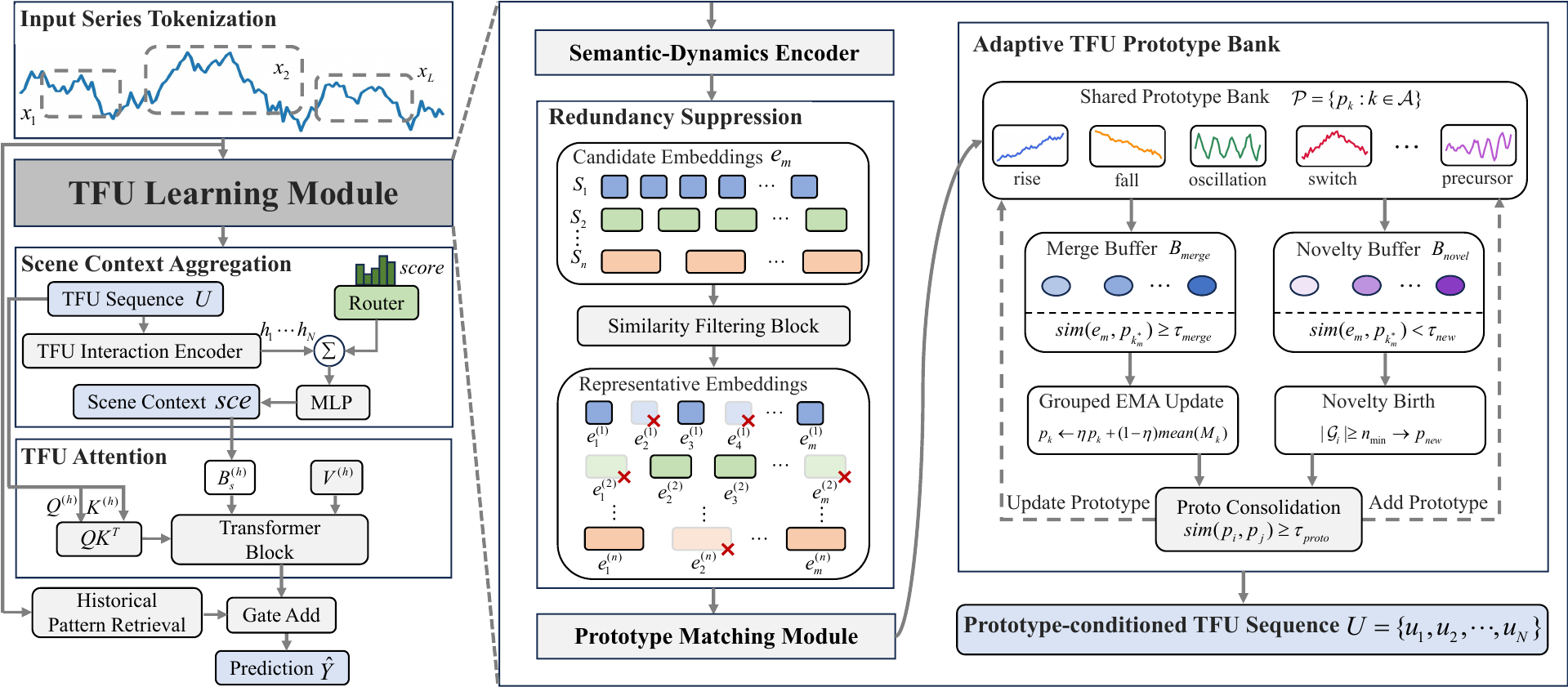}
    \caption{The overview of UniTFU. The TFU learning module encodes multi-scale temporal segments, suppresses redundant candidates, and organizes recurring local dynamics into a shared adaptive dictionary (TFU Prototype Bank) through prototype updating, novelty birth, and consolidation. The resulting TFU sequence and aggregated scene context guide temporal dynamics modeling, enabling explicit reuse and composition of shared local dynamics for cross-domain forecasting. The labeled curves (rise, fall, oscillation, switch, and precursor) are manually constructed illustrations, not semantic labels assigned to learned TFUs.}
    \label{fig:unitfu}
    \vspace{-3pt}
\end{figure}

We propose \textbf{UniTFU}, a unified time series foundation model that enables cross-domain forecasting through reusable Time Series Foundation Units (TFUs), as illustrated in Fig.~\ref{fig:unitfu}. UniTFU contains three core modules: (1) the \textbf{TFU Learning Module}, which discovers multi-scale local temporal structures and maps them to a shared TFU prototype dictionary; (2) \textbf{Scene Context Aggregation}, which summarizes the TFU sequence into a global context of local dynamic composition; and (3) \textbf{TFU Attention}, which incorporates the TFU sequence and scene context into a forecasting backbone. A lightweight predictor then produces the final forecast.

\noindent\textbf{Problem Statement.}
Given a historical observation sequence with $C$ variables,
$X=\{x_1,x_2,\ldots,x_L\}\in\mathbb{R}^{L\times C}$ over $L$ time steps,
UniTFU forecasts the subsequent $H$ steps as
$\hat{Y}_{L+1:L+H}=f_{\theta}(X)\in\mathbb{R}^{H\times C}$,
where $f_{\theta}$ denotes the time series foundation model.
UniTFU applies sliding windows at multiple temporal scales to extract candidate temporal segments, which are subsequently represented by reusable TFUs to provide structured context for forecasting.

\subsection{TFU Learning Module}
The TFU learning module discovers, reuses, and composes recurring local dynamics shared across domains, organizing them as reusable prototypes into a shared adaptive dictionary. It first encodes multi-scale temporal segments and suppresses redundant candidates, then matches the retained embeddings to shared TFU prototypes. The matched prototypes are aggregated at the token positions covered by the corresponding candidates to construct the prototype-conditioned TFU sequence. During training, prototype updating, novelty birth, and consolidation adapt the shared dictionary to newly observed local dynamics.
 
\noindent\textbf{Semantic Dynamics Encoder.}
Let ${Z_m} = [z_{m,1},z_{m,2}, \ldots ,z_{m,S}]\in\mathbb{R}^{{S}\times D}$ denote the $m$-th candidate segment in the token representation space extracted from consecutive tokens of a single variable, where $S$ denotes its temporal scale in tokens. Each token represents a temporal patch through a learned linear projection with positional information. UniTFU summarizes each segment by its global activation $\bar{z}_m$, element-wise local peak response $z_m^{\max}$, and temporal offset $\Delta z_m$,  which summarize average activations, peak responses, and changes between the first and last token representations. Each summary is a $D$-dimensional vector, yielding a $3D$-dimensional input to the segment encoder. These statistics are concatenated and projected by a lightweight encoder to obtain the segment embedding $e_m$:
\begin{equation}
\begin{gathered}
{{\bar z}_m} = \frac{1}{S}\sum\limits_{i = 1}^S {z_{m,i}} ,
\qquad
z_m^{\max } = {\max _{1\leq i\leq S}}z_{m,i},
\qquad
\Delta {z_m} = z_{m,S} - z_{m,1},\\
e_m=\mathrm{LayerNorm}\!\left(\mathrm{MLP}\!\left([\bar{z}_m;z_m^{\max};\Delta{z_m}]\right)\right)\in\mathbb{R}^{D}.
\end{gathered}
\end{equation}
The segment encoder can receive forecasting gradients through the similarity-dependent token-level composition described below, while prototype assignment is discrete.

\noindent\textbf{Redundancy Suppression.}
Overlapping windows may produce redundant candidates describing the same local temporal structure. Before prototype assignment, UniTFU filters candidates according to their temporal supports and representation similarity. For candidates extracted at temporal scale $S$, we denote the embedding of the $m$-th candidate as $e_m^{(S)}$. For two candidate segments $Z_i$ and $Z_j$ from the same variable and scale $S$, we compute the cosine similarity between their embeddings:
\begin{equation}
    \mathrm{sim}(e_i^{(S)},e_j^{(S)})
    =
    \frac{{(e_i^{(S)})^T} \cdot e_j^{(S)}}
    {
|e_i^{(S)}|_2
|e_j^{(S)}|_2
}.
\end{equation}
Within the same scale, a candidate is suppressed when its temporal support overlaps with another candidate and their similarity exceeds $\tau_{\mathrm{intra}}$. Across temporal scales, a shorter candidate covered by a longer one from the same variable is suppressed when their similarity exceeds $\tau_{\mathrm{cross}}$. This removes redundant candidates while preserving complementary local structures across temporal scales.

\noindent\textbf{Adaptive TFU Prototype Bank.}
After redundancy suppression, the retained candidate embeddings are assigned to a shared TFU prototype bank $\mathcal{P}=\{p_k:k\in\mathcal{A}\}$, where $\mathcal{A}\subseteq\{1,\ldots,K\}$ indexes active TFUs and $K$ is the maximum dictionary capacity. The bank starts empty, with the first retained candidate embedding initializing the first active TFU. Prototypes are maintained as non-gradient buffers rather than randomly initialized trainable parameters. For each retained embedding $e_m$, UniTFU selects the most similar prototype and records the corresponding matching confidence:
\begin{equation}
    k_m^*  = \mathop {\arg \max }_{k\in\mathcal{A}} {\rm{sim}}({e_m},{p_k}),\qquad {q_m} = {\rm{sim}}({e_m},{p_{k_m^* }}),
\end{equation}
where the index $k_m^*$ determines the TFU assigned to the candidate, while $q_m$ measures how well the candidate is explained by the selected prototype.

To construct a temporally aligned TFU sequence, the matched prototype is written back to the token positions covered by candidate $Z_m$. Let $\mathcal{I}_m$ denote the token positions covered by candidate $Z_m$ within its original variable. When multiple candidates cover the same position, their prototype representations are aggregated according to the matching confidence:
\begin{equation}
{u_t} = \frac{{\sum\limits_{{Z_m}:{\kern 1pt} t \in {{\cal I}_m}} {{w_m}} {p_{k_m^*}}}}{{\sum\limits_{{Z_m}:{\kern 1pt} t \in {{\cal I}_m}} {{w_m}} }},\qquad{w_m} = \exp \left( {\frac{{{q_m}}}{{\tau_{\mathrm{temp}}} }} \right),
\end{equation}
where $\tau_{\mathrm{temp}}$ is a temperature parameter. If only one candidate covers position $t$ and matches $p_k$, Eq.~(4) reduces to $u_t=p_k$. Positions not covered by any retained candidate preserve their original representations, yielding the prototype-conditioned TFU sequence $U = \{ {u_1},{u_2}, \cdot  \cdot  \cdot ,{u_N}\} $, where $N$ is the number of token positions in $U$. To adapt the TFU dictionary to newly observed dynamics, UniTFU periodically maintains the prototype bank using buffered candidate embeddings. Candidates well represented by existing prototypes are routed to the merge buffer $\mathcal{B}_{\mathrm{merge}}$, while insufficiently represented candidates are routed to the novelty buffer $\mathcal{B}_{\mathrm{novel}}$:
\begin{equation}
\begin{gathered}
    \mathrm{sim} ({e_m},p_{k_m^*})\geq\tau_{\mathrm{merge}}\Rightarrow e_m\in\mathcal{B}_{\mathrm{merge}},\\
    \mathrm{sim} ({e_m},p_{k_m^*})<\tau_{\mathrm{new}}\Rightarrow e_m\in\mathcal{B}_{\mathrm{novel}},
\end{gathered}
\end{equation}
where $\tau_{\mathrm{new}}<\tau_{\mathrm{merge}}$. Candidates with $\tau_{\mathrm{new}}\leq q_m<\tau_{\mathrm{merge}}$ are treated as uncertain and excluded from bank updating by default. Let $\mathcal{M}_k$ denote the merge candidates assigned to prototype $k$. UniTFU updates each prototype using a grouped exponential moving average:
\begin{equation}
    {p_k} \leftarrow \eta {p_k} + (1 - \eta )\frac{1}{{|{{\cal M}_k}|}}\sum\limits_{{e_m} \in {{\cal M}_k}} {{e_m}},
\end{equation}
where $\eta\in[0,1)$ controls the contribution of the previous prototype.
For existing dynamics, candidates in $\mathcal{B}_{\mathrm{merge}}$ refine matched prototypes through Eq.~(6), without changing the active count. For novel dynamics, each unused candidate $e_i\in\mathcal{B}_{\mathrm{novel}}$ serves as an anchor defining a similarity group $\mathcal{G}_i=\{e_j\in\mathcal{B}_{\mathrm{novel}}\mid \mathrm{sim}(e_i,e_j)\geq\tau_{\mathrm{novel}}\}$. If $|\mathcal{G}_i|\geq n_{\min}$ and capacity permits, its mean initializes a new prototype. Redundant prototypes satisfying $\mathrm{sim}(p_i,p_j)\geq\tau_{\mathrm{proto}}$ are consolidated. Thus, routing itself does not change the active count; prototype birth increases it, while consolidation decreases it, with maximum capacity $K$ fixed.
% where $\eta\in[0,1)$ controls the contribution of the previous prototype. The novelty buffer accumulates candidates that are insufficiently covered by the current bank. During periodic bank maintenance, each unused novelty candidate $e_i$ serves as an anchor and forms $\mathcal{G}_i=\{e_j\in\mathcal{B}_{\mathrm{novel}}\mid \mathrm{sim}(e_i,e_j)\geq\tau_{\mathrm{novel}}\}$. If $|\mathcal{G}_i|\geq n_{\min}$, the mean representation of $\mathcal{G}_i$ initializes a new prototype. Finally, prototypes satisfying $\mathrm{sim}(p_i,p_j)\geq\tau_{\mathrm{proto}}$ are consolidated to reduce dictionary redundancy. The dictionary therefore adapts both prototype representations and the number of active TFUs, while matched prototypes provide temporally aligned context for forecasting.

\subsection{Scene Context Aggregation}
The prototype-conditioned TFU sequence $U$ preserves token-level local temporal structures, while forecasting also requires a global representation of how these units compose within the current input window. UniTFU first applies a lightweight interaction encoder,
$\mathcal{H}=E(U)=[h_1,\ldots,h_N]$, where $h_t\in\mathbb{R}^{D}$, to contextualize the TFU representations. We then summarize the contextualized sequence through complementary attention, mean, and element-wise max pooling. Specifically, attention pooling uses scores $a_t=\mathrm{score}(h_t)$ with $\alpha_t=\mathrm{softmax}(a)_t$ to obtain $c_{\mathrm{attn}}=\sum_{t=1}^{N}\alpha_t h_t$, while $c_{\mathrm{mean}}=N^{-1}\sum_{t=1}^{N}h_t$ and $c_{\mathrm{max}}=\max_{1\leq t\leq N}h_t$ provide complementary global summaries. The resulting scene context is:
\begin{equation}
sce
    =
    \mathrm{LayerNorm}\!\left(
    \mathrm{MLP}_{\mathrm{scene}}\!\left(
    [c_{\mathrm{attn}};c_{\mathrm{mean}};c_{\mathrm{max}}]
    \right)\right)
    \in\mathbb{R}^{D}.
\end{equation}

\subsection{TFU Attention}
The TFU sequence $U$ and scene context $sce$ provide complementary
structural information for forecasting. UniTFU incorporates them into each token representation $z_i$ through gated residual conditioning:
\begin{equation}
\tilde{z}_i=\mathrm{LayerNorm}\!\left(
z_i+\gamma_u\phi_u(u_i)+\gamma_s\phi_s(sce)
\right),
\end{equation}
where $\phi_u$ and $\phi_s$ are learnable projections, and $\gamma_u$ and $\gamma_s$ are learnable scalar gates. The TFU-aware attention blocks then process the conditioned token sequence. For attention head $h$, let $q_i^{(h)}$ and $k_j^{(h)}$ denote the projected query and key from the conditioned sequence. We project $sce$ into a head-specific representation $s_h^{(h)}$
and compute $u_i^{(h)}=f_u^{(h)}([q_i^{(h)};s_h^{(h)}])$ and $v_j^{(h)}=f_v^{(h)}([k_j^{(h)};s_h^{(h)}])$. These $r$-dimensional representations define a low-rank scene-dependent relation bias that modulates the attention logits:
\begin{equation}
\begin{aligned}
B_s^{(h)}{(sce)_{ij}} = \frac{{(u_i^{{{(h)}}})^T v_j^{(h)}}}{{\sqrt r }},\qquad
{A^{(h)}} = {\rm{softmax}}\left( {\frac{{{Q^{(h)}}{K^{(h)}}^T + {\gamma _h}B_s^{(h)}(sce)}}{{\sqrt {{d_h}} }} + {M_{{\mathrm{p}}}}} \right).
\end{aligned}
\end{equation}
The head output is $O^{(h)}=A^{(h)}V^{(h)}$, where $d_h$ is the head dimension, $\gamma_h$ is a learnable head-specific gate, and $M_{\mathrm{p}}$ restricts token-level attention across positions, while TFU representations and scene context are constructed from the full observed input window.

\noindent\textbf{Prediction.}
The output of TFU Attention is mapped by a lightweight prediction head to obtain the dynamic forecast $Y_{\mathrm{d}}$. To exploit recurring historical patterns, UniTFU constructs $Y_{\mathrm{h}}$ by matching the latest query segment to historical segments within the current input window and reusing the immediately following segment of the matched candidate, with both segments required to lie entirely within the available history. The final prediction is $\hat{Y}=Y_{\mathrm{d}}+\lambda(Y_{\mathrm{h}}-Y_{\mathrm{d}})$, where $\lambda\in[0,1]$ is a learnable gate. More details are provided in Appendix~\ref{App_implementation}.

% To exploit recurring historical patterns, UniTFU further constructs a historical pattern forecast $Y_{\mathrm{h}}$ by retrieving the historical segment most similar to the latest observations and using its following segment. 

\section{EXPERIMENTS}
\label{sec:experiments}

We conduct extensive experiments to evaluate UniTFU. Section~\ref{sec:exp_settings} introduces the experimental settings. Section~\ref{exp_transfer} examines cross-domain generalization through controlled source-to-target transfer and target-domain adaptation, directly evaluating whether learned TFUs can be reused and updated across unseen domains. Section~\ref{exp_multi} evaluates standard supervised multivariate forecasting, while Section~\ref{exp_pre} compares UniTFU with publicly available pretrained TSFMs in zero-shot forecasting scenarios. Finally, Section~\ref{exp_ablat} and Section~\ref{exp_analysis} provide ablation studies and detailed model analyses.

\subsection{Experimental Settings}
\label{sec:exp_settings}

\noindent\textbf{Datasets \& Metrics.}
We evaluate UniTFU on seven forecasting datasets: the four ETT subsets, Electricity, Solar, and Weather, using mean squared error (MSE) and mean absolute error (MAE). Pre-training uses UTSD~\citep{liu2024timer} and LOTSA~\citep{woo2024unified}, excluding all target evaluation datasets to avoid data leakage.  Dataset details are provided in Appendix~\ref{APP_Datasets}.

\noindent\textbf{Baselines.}
We compare UniTFU with two categories of representative TSFMs: (1) TSFMs with general modeling paradigms, including GTM~\citep{chenggtm}, Timer-XL~\citep{liu2025timer}, and Timer~\citep{liu2024timer}; (2) large-scale pre-trained TSFMs, including Aurora~\citep{wu2025aurora}, Sundial~\citep{Sundial}, TiRex~\citep{auer2026tirex}, Time-MoE~\citep{shi2025time}, TimesFM~\citep{das2023decoder}, Chronos~\citep{ansari2024chronos}, and MOIRAI~\citep{woo2024unified}. For cross-domain forecasting, we additionally include UniTime~\citep{10.1145/3589334.3645434}, LangTime~\citep{3780338.3782211}, and LSTPO~\citep{11367281} as relevant transfer learning baselines. For supervised forecasting, we additionally compare with strong task-trained forecasting models, including PhaseFormer~\citep{niu2026phaseformer}, DLinear~\citep{zeng2023transformers}, PatchTST~\citep{PatchTST}, and iTransformer~\citep{iTransformer}. Detailed descriptions of the baselines are provided in Appendix~\ref{App_baselines}.

\noindent\textbf{Implementation Details.}
We use an input length of $L=672$. UniTFU uses input/output token lengths of 96, obtaining longer horizons through rolling prediction. Direct forecasting baselines follow the same rolling forecasting procedure. TFU learning uses scales ${S}=\{1,2,4\}$ and a maximum dictionary capacity of $K=128$ ($1024$ for pre-training), with the number of active TFUs determined adaptively. We set $\tau_{\mathrm{intra}}=\tau_{\mathrm{cross}}=0.92$, $\tau_{\mathrm{merge}}=0.82$, $\tau_{\mathrm{new}}=\tau_{\mathrm{merge}}-0.10$, $\tau_{\mathrm{novel}}=0.80$, and $\tau_{\mathrm{proto}}=0.97$. These settings are shared across datasets without dataset-specific tuning. Complete configurations and hyperparameter sensitivity analyses are provided in Appendix~\ref{App_implementation} and Fig.~\ref{fig:hyperparameter_sensitivity}.

\subsection{Cross-domain Time Series Forecasting} \label{exp_transfer}
We evaluate how effectively reusable TFUs transfer to unseen target domains under direct transfer and target-domain adaptation.

\noindent\textbf{Direct Transfer.}
We evaluate direct transfer across ETTh1, ETTm1, Weather, ECL, and Solar under the $672\rightarrow96$ setting. Each model is trained on one source domain and evaluated on the other four without target-domain training samples or parameter updates. We additionally include UniTime, LangTime, and LSTPO under the same transfer setting. We further evaluate two controlled variants: w/o TFU removes the TFU learning module, and w/ VQ replaces the adaptive TFU dictionary with a fixed-size VQ codebook~\citep{van2017neural} using cosine-similarity assignment and grouped EMA updates. Table~\ref{zero_transfer} averages results over four source domains per target. UniTFU achieves the lowest MSE on four of five targets and the lowest MAE on all five, with the best overall average errors, reducing overall MSE by 16.6\% and 12.8\% relative to GTM and Timer-XL, and by 13.4\% and 4.3\% relative to w/o TFU and w/ VQ. These results suggest that TFUs learned from source domains capture reusable local dynamics that remain effective on unseen target domains without dictionary adaptation.

\begin{table}[h]
\centering
\caption{Direct transfer results of representative TSFMs and cross-domain forecasting methods. Results are MSE/MAE averaged over four source domains. \textbf{\textcolor{red}{Red}}: the best, \textbf{\textcolor{blue}{Blue}}: the 2nd best. All the results are provided in Table~\ref{tab:direct_transfer_full} of Appendix~\ref{App_direct_transfer}.}
\label{zero_transfer}

\setlength{\tabcolsep}{3.5pt}
\renewcommand{\arraystretch}{0.92}

\resizebox{\textwidth}{!}{%
\begin{tabular}{c|cc|cc|cc|cc|cc|cc|cc|cc|cc}
\toprule
\multirow{2}{*}{Target Domain}
& \multicolumn{2}{c|}{UniTFU}
& \multicolumn{2}{c|}{w/ VQ}
& \multicolumn{2}{c|}{w/o TFU}
& \multicolumn{2}{c|}{GTM}
& \multicolumn{2}{c|}{Timer-XL}
& \multicolumn{2}{c|}{Timer}
& \multicolumn{2}{c|}{UniTime}
& \multicolumn{2}{c|}{LangTime}
& \multicolumn{2}{c}{LSTPO} \\
\cmidrule(lr){2-19}
& MSE & MAE & MSE & MAE & MSE & MAE
& MSE & MAE & MSE & MAE & MSE & MAE
& MSE & MAE & MSE & MAE & MSE & MAE \\
\midrule

ETTh1
& \textbf{\textcolor{red}{0.537}} & \textbf{\textcolor{red}{0.490}}
& \textbf{\textcolor{blue}{0.561}} & \textbf{\textcolor{blue}{0.501}}
& 0.633 & 0.527
& 0.653 & 0.577
& 0.635 & 0.541
& 0.635 & 0.533
& 0.608 & 0.544
& 0.597 & 0.539
& 0.607 & 0.528 \\

ETTm1
& 0.879 & \textbf{\textcolor{red}{0.601}}
& 0.919 & 0.617
& 1.097 & 0.661
& 0.947 & 0.645
& 0.894 & 0.614
& 1.104 & 0.669
& 0.875 & 0.621
& \textbf{\textcolor{blue}{0.856}} & 0.604
& \textbf{\textcolor{red}{0.855}} & \textbf{\textcolor{blue}{0.602}} \\

Weather
& \textbf{\textcolor{red}{0.245}} & \textbf{\textcolor{red}{0.287}}
& \textbf{\textcolor{blue}{0.250}} & \textbf{\textcolor{blue}{0.292}}
& 0.267 & 0.302
& 0.326 & 0.351
& 0.277 & 0.313
& 0.269 & 0.300
& 0.255 & 0.305
& 0.253 & 0.301
& \textbf{\textcolor{blue}{0.250}} & 0.302 \\

ECL
& \textbf{\textcolor{red}{0.579}} & \textbf{\textcolor{red}{0.560}}
& \textbf{\textcolor{blue}{0.601}} & \textbf{\textcolor{blue}{0.570}}
& 0.605 & 0.575
& 0.652 & 0.592
& 0.612 & 0.592
& 0.612 & 0.580
& 0.694 & 0.635
& 0.686 & 0.623
& 0.682 & 0.619 \\

Solar
& \textbf{\textcolor{red}{0.620}} & \textbf{\textcolor{red}{0.562}}
& \textbf{\textcolor{blue}{0.657}} & \textbf{\textcolor{blue}{0.590}}
& 0.700 & 0.600
& 0.851 & 0.653
& 0.862 & 0.691
& 0.706 & 0.602
& 0.885 & 0.753
& 0.867 & 0.734
& 0.861 & 0.748 \\

\midrule
Average
& \textbf{\textcolor{red}{0.572}} & \textbf{\textcolor{red}{0.500}}
& \textbf{\textcolor{blue}{0.598}} & \textbf{\textcolor{blue}{0.514}}
& 0.660 & 0.533
& 0.686 & 0.563
& 0.656 & 0.550
& 0.665 & 0.537
& 0.663 & 0.571
& 0.652 & 0.560
& 0.651 & 0.560 \\
\bottomrule
\end{tabular}%
}
\end{table}

\noindent\textbf{Target Domain Adaptation.}
We evaluate target-domain adaptation using 10\% of the target domain training samples under the $672\rightarrow96$ setting. Alongside standard 10\% few-shot results for Timer, Timer-XL, GTM, UniTime, LangTime, LSTPO, and UniTFU, we evaluate two UniTFU variants: Frozen-TFU updates only the forecasting head, while Updated-TFU additionally updates the TFU dictionary; both keep the remaining backbone fixed. As shown in Table~\ref{few_shot_transfer}, Updated-TFU improves average MSE and MAE over Frozen-TFU on all five target domains, reducing average MSE and MAE by 3.7\% and 3.9\%. These results indicate that source-domain TFUs provide a transferable prior for target-domain adaptation, while dictionary updates with limited target data further refine the reusable temporal structures. Updated-TFU also outperforms standard few-shot UniTFU on ETTm1, while remaining behind standard few-shot UniTFU and Timer-XL on ECL and Solar. This suggests that dictionary-level adaptation can be effective when target-domain dynamics can be accommodated by refining reusable TFUs, but dictionary updates do not always replace broader model adaptation.

\begin{table}[t]
\centering
\caption{Target-domain adaptation with 10\% training samples. Results are averaged over four source domains. \textbf{\textcolor{red}{Red}}: the best, \textbf{\textcolor{blue}{Blue}}: the 2nd best. Details are provided in Table~\ref{tab:target_adaptation_full} of Appendix~\ref{App_target_adaptation}.}
\label{few_shot_transfer}

\resizebox{\textwidth}{!}{%
\begin{tabular}{c|cc|cc|cc|cc|cc|cc|cc|cc|cc}
\toprule
\multirow{1}{*}{Type}
& \multicolumn{6}{c|}{10\% Few-shot}
& \multicolumn{2}{c|}{Frozen-TFU}
& \multicolumn{2}{c|}{Updated-TFU}
& \multicolumn{8}{c}{10\% Few-shot} \\
\midrule
\multirow{2}{*}{Target Domain}
& \multicolumn{2}{c|}{Timer}
& \multicolumn{2}{c|}{Timer-XL}
& \multicolumn{2}{c|}{GTM}
& \multicolumn{2}{c|}{UniTFU}
& \multicolumn{2}{c|}{UniTFU}
& \multicolumn{2}{c|}{UniTFU}
& \multicolumn{2}{c|}{UniTime}
& \multicolumn{2}{c|}{LangTime}
& \multicolumn{2}{c}{LSTPO} \\
\cmidrule(lr){2-19}
& MSE & MAE & MSE & MAE & MSE & MAE
& MSE & MAE & MSE & MAE & MSE & MAE
& MSE & MAE & MSE & MAE & MSE & MAE \\
\midrule

ETTh1
& 0.509 & 0.485
& 0.491 & 0.477
& 0.526 & 0.490
& 0.482 & 0.467
& 0.478 & \textbf{\textcolor{blue}{0.462}}
& \textbf{\textcolor{red}{0.447}} & \textbf{\textcolor{red}{0.453}}
& 0.479 & 0.473
& \textbf{\textcolor{blue}{0.470}} & 0.479
& 0.497 & 0.489 \\

ETTm1
& 0.575 & 0.507
& 0.570 & 0.483
& 0.625 & 0.487
& \textbf{\textcolor{blue}{0.447}} & \textbf{\textcolor{blue}{0.434}}
& \textbf{\textcolor{red}{0.440}} & \textbf{\textcolor{red}{0.429}}
& 0.555 & 0.474
& 0.486 & 0.450
& 0.520 & 0.456
& 0.469 & 0.437 \\

Weather
& \textbf{\textcolor{red}{0.176}} & \textbf{\textcolor{blue}{0.224}}
& \textbf{\textcolor{red}{0.176}} & 0.228
& 0.179 & 0.230
& 0.190 & 0.248
& 0.186 & 0.244
& \textbf{\textcolor{blue}{0.177}} & \textbf{\textcolor{red}{0.223}}
& 0.185 & 0.231
& 0.182 & 0.228
& 0.185 & 0.235 \\

ECL
& \textbf{\textcolor{blue}{0.166}} & 0.267
& 0.171 & \textbf{\textcolor{blue}{0.266}}
& 0.188 & 0.285
& 0.235 & 0.334
& 0.217 & 0.316
& \textbf{\textcolor{red}{0.164}} & \textbf{\textcolor{red}{0.259}}
& 0.172 & 0.274
& 0.171 & 0.268
& 0.176 & 0.281 \\

Solar
& \textbf{\textcolor{blue}{0.193}} & 0.254
& \textbf{\textcolor{red}{0.188}} & \textbf{\textcolor{blue}{0.240}}
& 0.204 & \textbf{\textcolor{red}{0.238}}
& 0.259 & 0.329
& 0.233 & 0.291
& 0.202 & 0.241
& 0.221 & 0.275
& 0.221 & 0.264
& 0.228 & 0.284 \\

\midrule
Average
& 0.323 & 0.347
& 0.319 & \textbf{\textcolor{blue}{0.339}}
& 0.344 & 0.346
& 0.322 & 0.362
& \textbf{\textcolor{blue}{0.310}} & 0.348
& \textbf{\textcolor{red}{0.309}} & \textbf{\textcolor{red}{0.330}}
& \textbf{\textcolor{red}{0.309}} & 0.341
& 0.313 & \textbf{\textcolor{blue}{0.339}}
& 0.311 & 0.345 \\
\bottomrule
\end{tabular}%
}
\end{table}

\subsection{MULTIVARIATE TIME SERIES FORECASTING} \label{exp_multi}

We follow iTransformer \citep{iTransformer} to evaluate UniTFU performance in standard supervised multivariate rolling forecasting. All models are trained from scratch on each dataset's training split and evaluated on its held-out test split. As shown in Table~\ref{multivariate_avg}, UniTFU achieves the lowest average MSE and MAE on five of seven datasets, with the lowest overall average MSE of 0.285 and MAE of 0.330. These results demonstrate the effectiveness of UniTFU in standard multivariate forecasting.

\begin{table}[h]
\centering
\caption{Multivariate forecasting results averaged over horizons $\{96,192,336,720\}$ with a lookback length of 672.  \textbf{\textcolor{red}{Red}}: the best, \textbf{\textcolor{blue}{Blue}}: the 2nd best. Details are provided in Table~\ref{tab:multivariate_full} of Appendix~\ref{App_multivariate}.}
\label{multivariate_avg}
\resizebox{\textwidth}{!}{
\begin{tabular}{c|cc|cc|cc|cc|cc|cc|cc|cc}

\toprule
\multirow{1}{*}{Type}
& \multicolumn{8}{c|}{TSFM}
& \multicolumn{8}{c}{Forecasting Model} \\
\midrule
\multirow{2}{*}{Models}
& \multicolumn{2}{c|}{\textbf{UniTFU}}
& \multicolumn{2}{c|}{GTM}
& \multicolumn{2}{c|}{Timer-XL}
& \multicolumn{2}{c|}{Timer}
& \multicolumn{2}{c|}{PhaseFormer}
& \multicolumn{2}{c|}{iTransformer}
& \multicolumn{2}{c|}{DLinear}
& \multicolumn{2}{c}{PatchTST} \\
& \multicolumn{2}{c|}{\textbf{(Ours)}}
& \multicolumn{2}{c|}{(2026)}
& \multicolumn{2}{c|}{(2025)}
& \multicolumn{2}{c|}{(2024)}
& \multicolumn{2}{c|}{(2026)}
& \multicolumn{2}{c|}{(2023)}
& \multicolumn{2}{c|}{(2023)}
& \multicolumn{2}{c}{(2022)} \\
\midrule
Metric 
& MSE & MAE 
& MSE & MAE 
& MSE & MAE 
& MSE & MAE 
& MSE & MAE 
& MSE & MAE 
& MSE & MAE 
& MSE & MAE \\
\midrule

ETTh1
& \textbf{\textcolor{red}{0.396}} & \textbf{\textcolor{red}{0.423}} 
& \textbf{\textcolor{blue}{0.398}} & 0.429 
& 0.409 & 0.433 
& 0.418 & 0.433 
& 0.418 & \textbf{\textcolor{blue}{0.427}} 
& 0.419 & 0.440 
& 0.436 & 0.451 
& 0.407 & 0.432 \\

ETTh2  
& \textbf{\textcolor{red}{0.366}} & \textbf{\textcolor{red}{0.405}} 
& 0.375 & 0.412 
& 0.376 & 0.413 
& 0.379 & 0.417 
& \textbf{\textcolor{blue}{0.369}} & 0.413 
& 0.395 & 0.424 
& 0.520 & 0.494 
& 0.372 & \textbf{\textcolor{blue}{0.409}} \\

ETTm1  
& \textbf{\textcolor{red}{0.337}} & \textbf{\textcolor{red}{0.371}} 
& \textbf{\textcolor{blue}{0.346}} & \textbf{\textcolor{blue}{0.374}} 
& 0.369 & 0.394 
& 0.347 & 0.378 
& 0.360 & 0.383 
& 0.387 & 0.409 
& 0.356 & 0.379 
& 0.363 & 0.394 \\

ETTm2  
& 0.285 & 0.336 
& 0.266 & 0.324 
& 0.276 & 0.330 
& \textbf{\textcolor{red}{0.260}} & \textbf{\textcolor{red}{0.318}} 
& 0.271 & 0.326 
& 0.296 & 0.344 
& 0.284 & 0.346 
& \textbf{\textcolor{blue}{0.263}} & \textbf{\textcolor{blue}{0.323}} \\

ECL  
& \textbf{\textcolor{red}{0.164}} & \textbf{\textcolor{red}{0.259}} 
& 0.173 & 0.272 
& 0.169 & 0.262 
& 0.169 & 0.261 
& 0.168 & \textbf{\textcolor{blue}{0.260}} 
& \textbf{\textcolor{blue}{0.166}} & \textbf{\textcolor{red}{0.259}} 
& \textbf{\textcolor{blue}{0.166}} & 0.261 
& 0.173 & 0.272 \\

Weather  
& \textbf{\textcolor{blue}{0.235}} & \textbf{\textcolor{red}{0.270}} 
& 0.240 & 0.274 
& 0.242 & 0.279 
& 0.243 & 0.276 
& \textbf{\textcolor{blue}{0.235}} & 0.278 
& 0.239 & 0.278 
& 0.240 & 0.291 
& \textbf{\textcolor{red}{0.231}} & \textbf{\textcolor{blue}{0.271}} \\

Solar  
& \textbf{\textcolor{red}{0.209}} & \textbf{\textcolor{red}{0.250}} 
& 0.231 & \textbf{\textcolor{red}{0.250}} 
& 0.219 & \textbf{\textcolor{blue}{0.262}} 
& 0.220 & 0.266 
& 0.216 & 0.273 
& 0.219 & 0.290 
& 0.221 & 0.282 
& \textbf{\textcolor{blue}{0.210}} & 0.294 \\
\bottomrule
\end{tabular}
}
\end{table}

\begin{table}[h]
\centering
\caption{Pretrained forecasting results averaged over horizons $\{96,192,336,720\}$ with a lookback length of 672. ($'-'$) indicates that the dataset was included in pre-training and therefore excluded from evaluation. \textbf{\textcolor{red}{Red}}: the best, \textbf{\textcolor{blue}{Blue}}: the 2nd best. Details are provided in Table~\ref{tab:pretrained_full} of Appendix~\ref{App_pretrained}.}

\label{pretrained_avg}
\resizebox{\textwidth}{!}{
\begin{tabular}{c|cc|cc|cc|cc|cc|cc|cc|cc}

\toprule
\multirow{2}{*}{Models}
& \multicolumn{2}{c|}{\textbf{UniTFU}}
& \multicolumn{2}{c|}{Aurora}
& \multicolumn{2}{c|}{Sundial}
& \multicolumn{2}{c|}{TiRex}
& \multicolumn{2}{c|}{Time-MoE}
& \multicolumn{2}{c|}{MOIRAI}
& \multicolumn{2}{c|}{TimesFM}
& \multicolumn{2}{c}{Chronos} \\
& \multicolumn{2}{c|}{\textbf{(Ours)}}
& \multicolumn{2}{c|}{(2026)}
& \multicolumn{2}{c|}{(2025)}
& \multicolumn{2}{c|}{(2025)}
& \multicolumn{2}{c|}{(2025)}
& \multicolumn{2}{c|}{(2024)}
& \multicolumn{2}{c|}{(2023)}
& \multicolumn{2}{c}{(2024)} \\
\midrule
Metric 
& MSE & MAE 
& MSE & MAE 
& MSE & MAE 
& MSE & MAE 
& MSE & MAE 
& MSE & MAE 
& MSE & MAE 
& MSE & MAE \\
\midrule

ETTh1
& \textbf{\textcolor{red}{0.380}} & \textbf{\textcolor{red}{0.403}}
& 0.387 & 0.418
& \textbf{\textcolor{blue}{0.382}} & \textbf{\textcolor{blue}{0.406}}
& 0.437 & 0.417
& 0.450 & 0.434
& 0.451 & 0.441
& 0.413 & 0.408
& 0.579 & 0.473 \\

ETTh2
& 0.348 & \textbf{\textcolor{blue}{0.379}}
& 0.347 & 0.389
& \textbf{\textcolor{blue}{0.346}} & 0.383
& \textbf{\textcolor{red}{0.343}} & \textbf{\textcolor{red}{0.369}}
& 0.368 & 0.395
& 0.405 & 0.412
& 0.370 & 0.388
& 0.444 & 0.429 \\

ETTm1
& \textbf{\textcolor{red}{0.361}} & \textbf{\textcolor{red}{0.370}}
& 0.424 & 0.427
& \textbf{\textcolor{blue}{0.363}} & \textbf{\textcolor{blue}{0.372}}
& 0.409 & 0.374
& 0.420 & 0.388
& 0.488 & 0.463
& 0.384 & \textbf{\textcolor{red}{0.370}}
& 0.644 & 0.498 \\

ETTm2
& \textbf{\textcolor{red}{0.272}} & \textbf{\textcolor{red}{0.314}}
& 0.297 & 0.344
& \textbf{\textcolor{blue}{0.274}} & 0.320
& 0.291 & \textbf{\textcolor{blue}{0.316}}
& 0.307 & 0.335
& 0.362 & 0.366
& 0.293 & 0.317
& 0.372 & 0.380 \\

ECL
& 0.158 & 0.245
& 0.179 & 0.276
& \textbf{\textcolor{red}{0.145}} & \textbf{\textcolor{blue}{0.242}}
& \textbf{\textcolor{blue}{0.156}} & \textbf{\textcolor{red}{0.240}}
& $-$ & $-$
& 0.211 & 0.303
& $-$ & $-$
& 0.263 & 0.319 \\

Weather
& \textbf{\textcolor{red}{0.240}} & \textbf{\textcolor{red}{0.258}}
& 0.275 & 0.295
& \textbf{\textcolor{blue}{0.242}} & 0.273
& 0.244 & \textbf{\textcolor{blue}{0.260}}
& 0.260 & 0.287
& 0.353 & 0.320
& $-$ & $-$
& 0.324 & 0.312 \\

Solar
& \textbf{\textcolor{red}{0.201}} & 0.252
& 0.216 & 0.255
& \textbf{\textcolor{blue}{0.203}} & \textbf{\textcolor{blue}{0.240}}
& 0.335 & 0.274
& 0.348 & 0.287
& 0.719 & 0.702
& 0.259 & \textbf{\textcolor{red}{0.238}}
& 0.719 & 0.606 \\

\bottomrule
\end{tabular}
}

\end{table}

\subsection{PRETRAINED TIME SERIES FORECASTING} \label{exp_pre}
We compare pretrained UniTFU with seven pretrained TSFMs using a 672-step lookback and horizons $\{96,192,336,720\}$. For UniTFU, all evaluation datasets are excluded from pre-training, while competing TSFMs are evaluated only on datasets not reported in their pre-training corpora. UniTFU is evaluated zero-shot using a single pretrained checkpoint, without target-domain adaptation. As shown in Table~\ref{pretrained_avg}, UniTFU achieves the lowest average MSE on five of seven datasets and the lowest average MAE on four, with an overall MSE of 0.280 and MAE of 0.317. These results support the effectiveness of reusable TFUs for transferring shared local dynamics across downstream datasets.

\subsection{ABLATION STUDY} \label{exp_ablat}

We conduct ablation experiments under the $672\rightarrow96$ forecasting setting using four random seeds, including the benchmark seed 2021. We consider five variants: w/o TFU and w/ VQ are defined in Section~\ref{exp_transfer}; w/o Sce excludes scene context aggregation; w/o Att replaces TFU attention with standard self-attention; w/o Res removes the historical-pattern residual. As shown in Table~\ref{ablation}, removing individual components generally increases mean errors, supporting their complementary contributions. The w/o TFU results support the contribution of TFU learning, while the comparison with w/ VQ supports the adaptive dictionary design.

\begin{table}[h]
\centering
\caption{Ablation results under the $672\rightarrow96$ forecasting setting. For ETT, mean and std are averaged over ETTh1, ETTh2, ETTm1, and ETTm2. All the results are provided in Table~\ref{tab:ablation_full}.}
\label{ablation}
\begingroup
\newcommand{\abcell}[2]{%
\makecell[c]{$#1$\\[-1.5pt]$\pm\,#2$}}
\newcommand{\abest}[2]{%
\makecell[c]{$\mathbf{#1}$\\[-1.5pt]$\bm{\pm\,#2}$}}
\setlength{\tabcolsep}{3pt}
\renewcommand{\arraystretch}{1.05}
\resizebox{\textwidth}{!}{%
\begin{tabular}{l|cc|cc|cc|cc|cc|cc}
\toprule
Models
& \multicolumn{2}{c|}{UniTFU}
& \multicolumn{2}{c|}{w/o TFU}
& \multicolumn{2}{c|}{w/o Sce}
& \multicolumn{2}{c|}{w/o Att}
& \multicolumn{2}{c|}{w/o Res}
& \multicolumn{2}{c}{w/ VQ} \\
\midrule
Metric
& MSE & MAE & MSE & MAE & MSE & MAE
& MSE & MAE & MSE & MAE & MSE & MAE \\
\midrule

ETT
& \abest{0.279}{0.0026}
& \abest{0.336}{0.0028}
& \abcell{0.287}{0.0023}
& \abcell{0.342}{0.0025}
& \abcell{0.283}{0.0026}
& \abcell{0.339}{0.0028}
& \abcell{0.283}{0.0046}
& \abcell{0.339}{0.0045}
& \abcell{0.285}{0.0034}
& \abcell{0.341}{0.0034}
& \abcell{0.284}{0.0022}
& \abcell{0.340}{0.0027} \\
\midrule

ECL
& \abest{0.133}{0.0034}
& \abest{0.231}{0.0051}
& \abcell{0.138}{0.0019}
& \abcell{0.236}{0.0021}
& \abcell{0.134}{0.0026}
& \abcell{0.232}{0.0028}
& \abcell{0.138}{0.0033}
& \abcell{0.236}{0.0037}
& \abcell{0.136}{0.0028}
& \abcell{0.234}{0.0037}
& \abcell{0.136}{0.0013}
& \abcell{0.233}{0.0019} \\
\midrule

Weather
& \abcell{0.159}{0.0030}
& \abest{0.208}{0.0028}
& \abcell{0.162}{0.0033}
& \abcell{0.214}{0.0034}
& \abcell{0.161}{0.0028}
& \abcell{0.210}{0.0037}
& \abest{0.157}{0.0028}
& \abcell{0.209}{0.0024}
& \abcell{0.158}{0.0029}
& \abcell{0.210}{0.0030}
& \abcell{0.161}{0.0005}
& \abcell{0.212}{0.0010} \\
\midrule

Solar
& \abest{0.182}{0.0024}
& \abest{0.221}{0.0024}
& \abcell{0.192}{0.0022}
& \abcell{0.227}{0.0022}
& \abcell{0.195}{0.0025}
& \abcell{0.231}{0.0022}
& \abcell{0.198}{0.0036}
& \abcell{0.233}{0.0037}
& \abcell{0.188}{0.0028}
& \abcell{0.224}{0.0029}
& \abcell{0.187}{0.0038}
& \abcell{0.224}{0.0018} \\
\bottomrule
\end{tabular}%
}
\endgroup
\end{table}

% relative to the evaluated VQ replacement. 

\subsection{Model Analysis}
\label{exp_analysis}

% \begin{table}[t]
% \centering
% \caption{
% Efficiency analysis under the $672\rightarrow336$ setting
% with batch size 1.
% }
% \label{tab:efficiency}
% \setlength{\tabcolsep}{3.5pt}
% \renewcommand{\arraystretch}{0.92}
% \resizebox{0.9\textwidth}{!}{%
% \begin{tabular}{l|ccc|ccc}
% \toprule
% Models
% & \multicolumn{3}{c|}{Weather}
% & \multicolumn{3}{c}{ECL} \\
% \cmidrule(lr){2-4}\cmidrule(lr){5-7}
% & Params (M) & Latency (ms) & Memory (MB)
% & Params (M) & Latency (ms) & Memory (MB) \\
% \midrule
% Timer
% & 12.709 & 29.140 & 76.8
% & 6.404 & 41.057 & 108.8 \\
% Timer-XL
% & 12.710 & 100.976 & 65.4
% & 6.405 & 156.321 & 880.6 \\
% GTM
% & 35.996 & 85.196 & 187.7
% & 33.015 & 174.764 & 691.9 \\
% UniTFU
% & 119.137 & 402.177 & 554.2
% & 119.137 & 678.949 & 1448.4 \\
% \bottomrule
% \end{tabular}%
% }
% \end{table}

% Table~\ref{tab:efficiency} shows that the evaluated UniTFU configuration uses more parameters and incurs higher inference latency and peak GPU memory than the baselines. 

\begin{figure*}[t]
    \centering
    \begin{subfigure}[t]{0.46\textwidth}
        \centering
        \includegraphics[width=\linewidth]
        {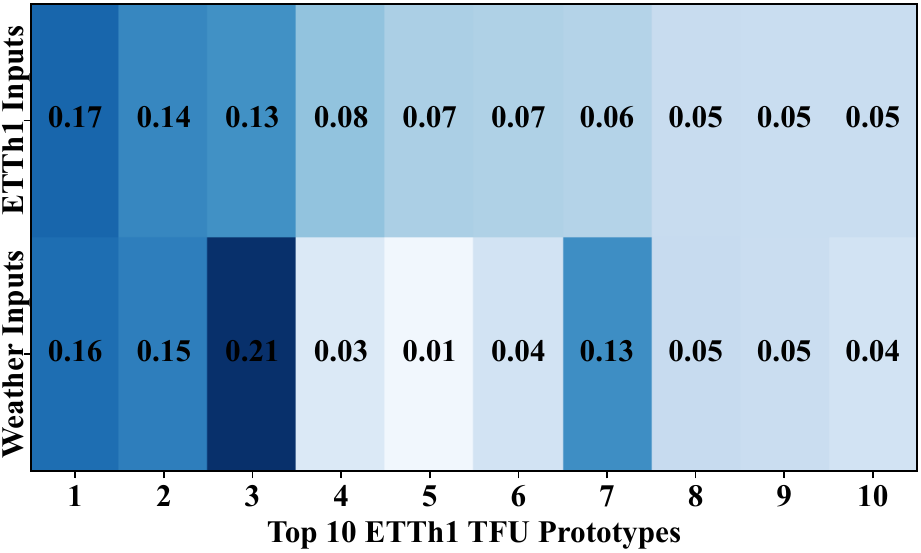}
        \caption{Cross-domain TFU reuse.}
        \label{fig:tfu_reuse}
    \end{subfigure}
    \hfill
    \begin{subfigure}[t]{0.48\textwidth}
        \centering
        \includegraphics[width=\linewidth]
        {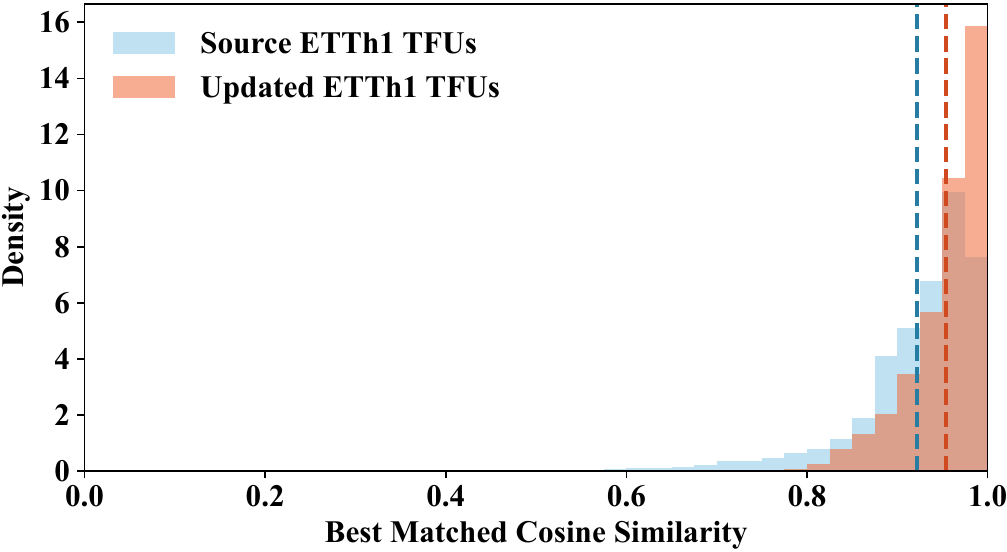}
        \caption{Target-domain TFU refinement.}
        \label{fig:tfu_refinement}
    \end{subfigure}
    \caption{
    TFU analysis from ETTh1 to Weather:
    (a) prototype usage in both domains;
    (b) best-match cosine similarities of Weather candidates
    before and after dictionary updating.
    }
    \label{fig:tfu_transfer_analysis}
    \vspace{-3pt}
\end{figure*}

\noindent\textbf{Cross-domain TFU Reuse Analysis.}
We further analyze TFU reuse, \begin{wrapfigure}{r}{0.48\textwidth}
    \centering
    \includegraphics[width=\linewidth]
    {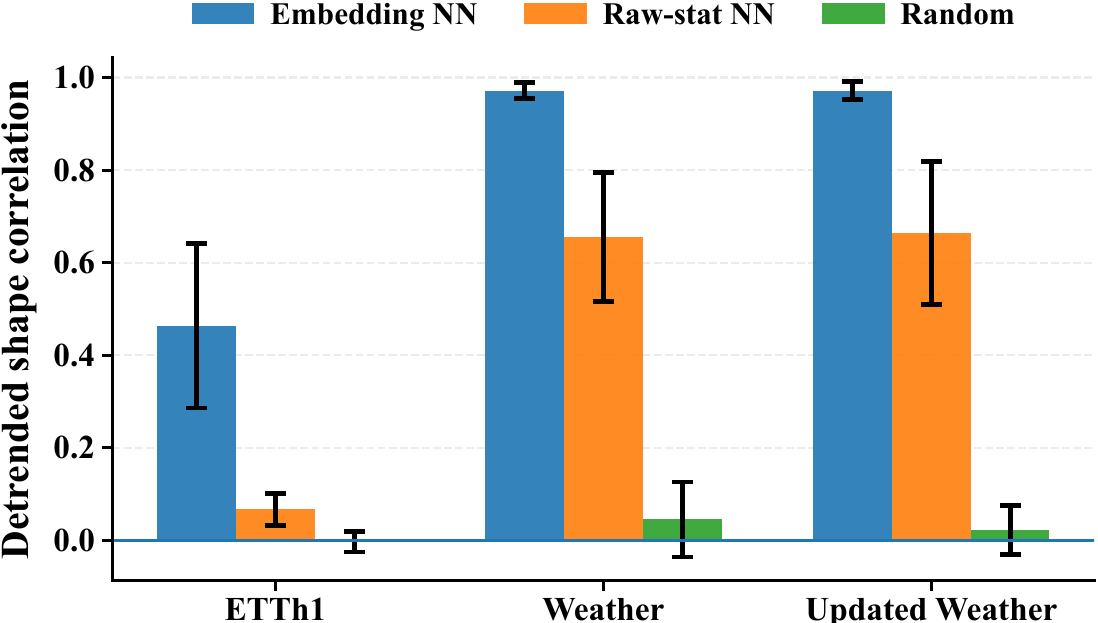}
    \caption{Local shape correlation across retrieval strategies.}
    \label{fig:embedding_shape}
\end{wrapfigure} adaptation, and local shape representations from ETTh1 to Weather. Fig.~\ref{fig:tfu_reuse} shows that Weather inputs reuse TFUs frequently activated in ETTh1. As shown in Fig.~\ref{fig:tfu_refinement}, updating the dictionary with limited Weather data increases best-match cosine similarities, indicating closer matches between target-domain candidates and the updated TFUs. We examine the local waveform information retained by the segment encoder through nearest-neighbor retrieval, using raw temporal statistics and random retrieval as baselines. Self-matches and overlapping segments are excluded, and waveform correlation is computed after linear detrending and normalization. Fig.~\ref{fig:embedding_shape} shows that embedding-based retrieval consistently identifies segments with higher shape correlations across all three settings. Together, these analyses support cross-domain TFU reuse and adaptation, with retrieval results further supporting local shape information beyond coarse amplitude and linear trend statistics. We extend the analysis to four source--target pairs, where embedding-based retrieval achieves the highest local shape correlation in 7 of 8 frozen/updated settings. Across 20 transfers, prototype usage is negatively correlated with window-level forecasting MSE in 17 settings, while similarity gains are associated with MSE reductions (Spearman $\rho=0.721$). However, higher matching similarity does not always improve forecasting accuracy. Complete results are provided in Appendix~\ref{App_model_analysis}.

\begin{table}[h]
\centering
\caption{Efficiency analysis under the $672\rightarrow336$ setting with batch size 1.}
\label{tab:efficiency}
\setlength{\tabcolsep}{3.5pt}
\renewcommand{\arraystretch}{0.92}
\resizebox{0.7\textwidth}{!}{%
\begin{tabular}{l|c|cc|cc}
\toprule
\multirow{2}{*}{Models}
& \multirow{2}{*}{Params (M)}
& \multicolumn{2}{c|}{Weather}
& \multicolumn{2}{c}{ECL} \\
\cmidrule(lr){3-4}\cmidrule(lr){5-6}
& & Latency (ms) & Memory (MB)
& Latency (ms) & Memory (MB) \\
\midrule
Timer
& 12.709 / 6.404
& 29.140 & 76.8
& 41.057 & 108.8 \\
Timer-XL
& 12.710 / 6.405
& 100.976 & 65.4
& 156.321 & 880.6 \\
GTM
& 35.996 / 33.015
& 85.196 & 187.7
& 174.764 & 691.9 \\
DLinear
& 0.129
& 0.962 & 9.3
& 1.398 & 13.2 \\
PatchTST
& 1.432
& 8.964 & 50.7
& 36.523 & 528.8 \\
iTransformer
& 4.600
& 6.306 & 28.5
& 10.437 & 43.1 \\
PhaseFormer
& 0.007
& 15.491 & 10.1
& 16.921 & 17.1 \\
UniTFU
& 119.137
& 402.177 & 554.2
& 678.949 & 1448.4 \\
\bottomrule
\end{tabular}%
}
\end{table}

\noindent\textbf{Inference Efficiency.}
Table~\ref{tab:efficiency} shows that the evaluated UniTFU configuration uses more parameters and incurs higher inference latency and peak GPU memory than the baselines. Under the same inference setting, UniTFU achieves only marginal supervised forecasting gains over lightweight baselines, while incurring substantially higher computational costs. On Weather, PatchTST achieves lower MSE. This trade-off reflects UniTFU's emphasis on cross-domain reusability rather than inference efficiency.

\begin{figure*}[h]
    \centering
    \includegraphics[width=0.95\textwidth]{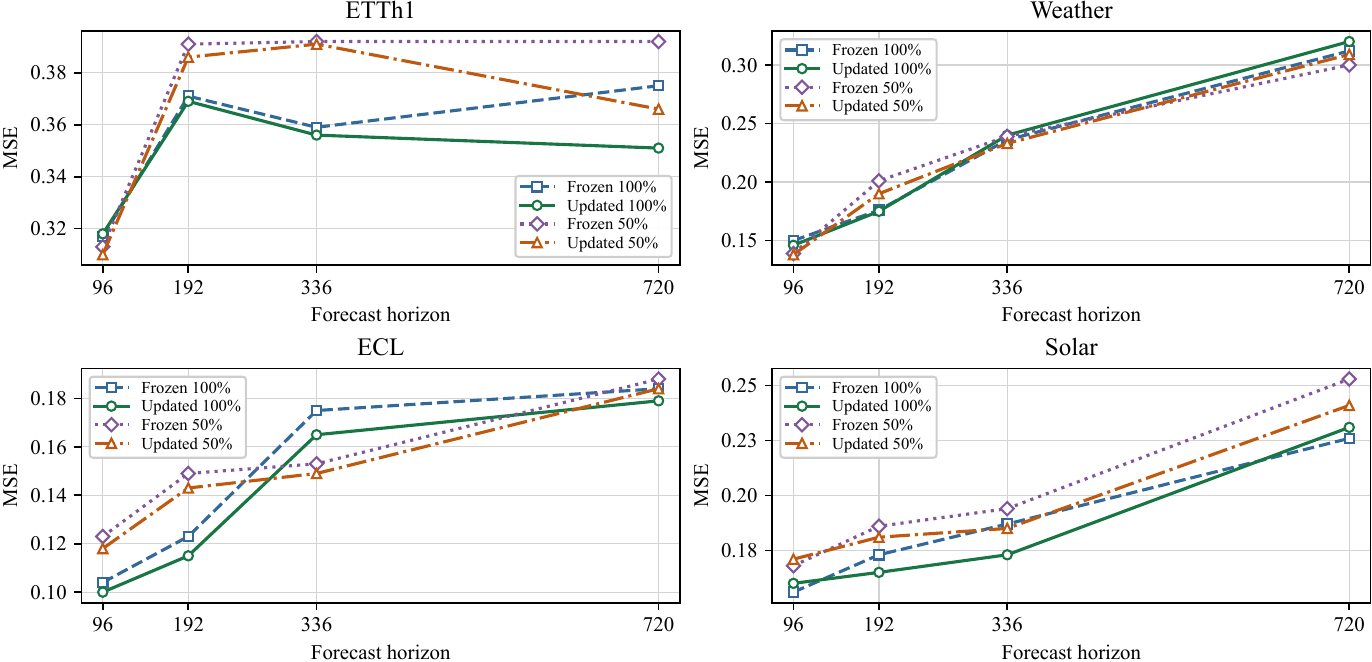}
    \caption{Effect of TFU dictionary adaptation after large-scale pretraining.}
    \label{fig:dictionary_adaptation}
    % \vspace{-3pt}
\end{figure*}

\noindent\textbf{Dictionary Adaptation after Large-scale Pretraining.}
We investigate TFU dictionary adaptation using the pretrained UniTFU checkpoint from Section~\ref{exp_pre}. We retain 100\% or 50\% of its active prototypes and compare frozen and updated dictionaries using the same 10\% target training data, with the backbone fixed. As shown in Fig.~\ref{fig:dictionary_adaptation}, dictionary updates reduce average MSE by 1.68\% and 2.27\% with 100\% and 50\% initial prototype retention, respectively. Although the gains vary across datasets and forecasting horizons, these results indicate that dictionary updates remain beneficial on average after large-scale pretraining, with different gains across initial vocabulary sizes.

\section{Conclusion}
We propose UniTFU, a time series foundation model that organizes recurring local dynamics into reusable and adaptable TFUs for cross-domain forecasting. Within UniTFU, comparisons with w/o TFU and a fixed-size EMA codebook support the contributions of TFU learning to cross-domain forecasting. Model analyses further support the retention of local shape information beyond coarse temporal statistics. These findings suggest that TFU updates improve average target-domain adaptation. However, TFU benefits vary across transfer settings, and the full model incurs higher inference costs than the evaluated baselines. Future work will investigate the adaptation of TFUs to a broader range of downstream time series tasks and reduce the computational cost of TFU-based forecasting.

\bibliography{iclr2027_conference}
\bibliographystyle{iclr2027_conference}

\clearpage
\appendix
\section{Experimental Details}

\subsection{Datasets} \label{APP_Datasets}

\noindent\textbf{Forecasting Datasets.}
We evaluate UniTFU on seven widely used real-world forecasting benchmarks: ETTh1, ETTh2, ETTm1, ETTm2, Electricity (ECL), Weather, and Solar. These benchmarks differ in application domain, sampling frequency, and number of variables, enabling evaluation of forecasting performance and cross-domain generalization under heterogeneous temporal dynamics. The four ETT datasets contain seven electricity transformer variables, sampled hourly in ETTh1 and ETTh2 and every 15 minutes in ETTm1 and ETTm2. ECL records hourly electricity consumption from 321 clients. Weather contains 21 meteorological variables sampled every 10 minutes, while Solar records power generation from 137 photovoltaic plants at the same sampling frequency. Following the standard evaluation setting, we partition each dataset chronologically into training, validation, and test sets. The ETT datasets use a 12/4/4-month split, while ECL, Weather, and Solar use a $7:1:2$ split. Table~\ref{tab:dataset_statistics} summarizes the dataset statistics.

\begin{table}[h]
\centering
\caption{Statistics of the forecasting benchmarks.}
\label{tab:dataset_statistics}
\begin{tabular}{lcccc}
\toprule
Dataset & Domain & Variables & Frequency & Time Points \\
\midrule
ETTh1        & Electricity & 7   & 1 hour & 17,420 \\
ETTh2        & Electricity & 7   & 1 hour & 17,420 \\
ETTm1        & Electricity & 7   & 15 min & 69,680 \\
ETTm2        & Electricity & 7   & 15 min & 69,680 \\
Electricity  & Electricity & 321 & 1 hour & 26,304 \\
Weather      & Climate     & 21  & 10 min & 52,696 \\
Solar & Energy      & 137 & 10 min & 52,560 \\
\bottomrule
\end{tabular}
\end{table}

\noindent\textbf{Pre-training Datasets.}
We construct the pre-training corpus from UTSD~\citep{liu2024timer} and LOTSA~\citep{woo2024unified}. UTSD contains approximately one billion time points across seven domains, while LOTSA comprises over 27 billion observations across nine domains. Together, these collections span diverse application domains, sampling frequencies, and temporal dynamics, providing a heterogeneous corpus for learning reusable local temporal structures. To ensure a strict evaluation of generalization and avoid data leakage, all datasets used as target evaluation benchmarks are excluded from the pre-training corpus. This exclusion applies to UniTFU's pre-training corpus; the pre-training exposure of competing pretrained TSFMs follows the datasets documented in their original papers and released model descriptions.

\subsection{Baselines} \label{App_baselines}

We compare UniTFU with representative TSFMs and supervised forecasting models. According to their roles in our experiments, we organize the baselines into four groups: TSFMs with general modeling paradigms, cross-domain forecasting methods, large-scale pre-trained TSFMs evaluated using released checkpoints, and task-trained forecasting models. This grouping reflects their experimental roles rather than mutually exclusive model categories.

\noindent\textbf{TSFMs with general modeling paradigms.}
GTM~\citep{chenggtm} uses Fourier attention to learn temporal-granularity-aware representations and combines reconstruction and autoregressive objectives through a unified pre-training strategy. Timer-XL~\citep{liu2025timer} extends next-token prediction to multivariate time series and introduces TimeAttention to capture intra-series and inter-series dependencies within a decoder-only Transformer. Timer~\citep{liu2024timer} represents time series as non-overlapping patch tokens and performs autoregressive forecasting through next-token prediction. These models serve as baselines for evaluating unified forecasting and controlled cross-domain generalization.

\noindent\textbf{Cross-domain forecasting methods.}
UniTime~\citep{10.1145/3589334.3645434} incorporates domain instructions into a unified language-based forecasting model. LangTime~\citep{3780338.3782211} uses temporal comprehension prompts and reinforcement learning to improve cross-domain forecasting. LSTPO~\citep{11367281} models temporal dependency preferences through meta-learning to facilitate cross-domain transfer and adaptation.

\noindent\textbf{Large-scale pre-trained TSFMs.}
Aurora~\citep{wu2025aurora} integrates time series with multimodal information for generative forecasting across heterogeneous domains. Sundial~\citep{Sundial} models the conditional distribution of future patches in continuous space using TimeFlow Loss, supporting probabilistic zero-shot forecasting. TiRex~\citep{auer2026tirex} is an xLSTM-based pre-trained forecasting model that uses contiguous patch masking to strengthen state tracking and support probabilistic forecasting over short and long horizons. Time-MoE~\citep{shi2025time} combines autoregressive forecasting with a sparse mixture-of-experts architecture to scale model capacity through selective expert activation. TimesFM~\citep{das2023decoder} is a patch-based decoder-only model pre-trained for zero-shot time series forecasting. Chronos~\citep{ansari2024chronos} scales and quantizes continuous observations into discrete tokens and trains a language model to generate probabilistic forecasts. MOIRAI~\citep{woo2024unified} accommodates heterogeneous sampling frequencies and variable dimensions through multiple patch sizes and any-variate attention, and uses a mixture distribution for probabilistic forecasting.

\noindent\textbf{Task-trained forecasting models.}
PhaseFormer~\citep{niu2026phaseformer} uses compact phase representations and lightweight cross-phase routing for efficient forecasting. DLinear~\citep{zeng2023transformers} decomposes time series into trend and remainder components and predicts them using separate linear mappings. PatchTST~\citep{PatchTST} combines temporal patching with a channel-independent Transformer to model dependencies within each variable. iTransformer~\citep{iTransformer} embeds the historical observations of each variable into a variate token and applies self-attention across variables to capture multivariate dependencies.

For reproduced baselines, we use the official implementations and follow the recommended configurations where available. All models in the supervised forecasting experiments follow the same standard rolling forecasting setting as UniTFU, using identical chronological data partitions and evaluation horizons. In controlled cross-domain experiments, all compared models follow the same 20 source-target pairs under the $672\rightarrow96$ setting. Direct transfer uses no target training data, while target-domain adaptation fine-tunes source-trained models using 10\% of the target training split. For pre-trained forecasting, we evaluate all competing TSFMs using their publicly released checkpoints. The lookback length is fixed to 672 for all models, and predictions are evaluated at horizons $\{96,192,336,720\}$ using MSE and MAE on the same target test sets. Because the compared TSFMs are pre-trained on different corpora, pre-training overlap is determined according to the datasets documented in their original papers or released model descriptions. This comparison evaluates released pretrained models rather than controlling for pre-training data and compute.

\subsection{Implementation Details}
\label{App_implementation}

We implement UniTFU in PyTorch and conduct experiments on NVIDIA RTX 4090 GPUs. Table~\ref{tab:implementation_details} summarizes the default model and training configurations. Unless explicitly varied in ablation or sensitivity experiments, the TFU-specific hyperparameters are shared across datasets without dataset-specific tuning. The prototype capacity is increased from 128 to 1024 for large-scale pre-training.

\begin{table}[t]
\centering
\caption{Implementation settings of UniTFU. TFU-specific hyperparameters are shared across datasets unless explicitly varied in ablation or sensitivity experiments. The prototype capacity is 128 for non-pretrained experiments and 1024 for pre-training.}
\label{tab:implementation_details}
\small
\begin{tabular}{lll}
\toprule
Category & Hyperparameter & Value \\
\midrule
\multirow{4}{*}{Forecasting}
& Input length $L$ & 672 \\
& Input token length & 96 \\
& Output token length & 96 \\
& Supervised prediction horizons & $\{96,192,336,720\}$ \\
\midrule
\multirow{4}{*}{Backbone}
& Number of layers & 2 \\
& Hidden dimension $D$ & 1536 \\
& FFN dimension & 3172 \\
& Number of attention heads & 8 \\
\midrule
\multirow{4}{*}{Scene Context}
& Interaction encoder layers & 1 \\
& Hidden dimension & 1536 \\
& Number of attention heads & 4 \\
& Scene-relation rank $r$ & 1 \\
\midrule
\multirow{8}{*}{Optimization}
& Optimizer & Adam \\
& Standard initial learning rate & $1\times10^{-4}$ \\
& Pre-training initial learning rate & $2\times10^{-5}$ \\
& Pre-training effective global batch size & 4096 \\
& Maximum training epochs & 10 \\
& Forecasting objective & MSE \\
& Learning-rate schedule & Cosine annealing \\
& Early-stopping patience & 10 epochs \\
\midrule
\multirow{12}{*}{TFU Learning}
& Temporal scales ${S}$ & $\{1,2,4\}$ tokens \\
& $\tau_{\mathrm{intra}}$ & 0.92 \\
& $\tau_{\mathrm{cross}}$ & 0.92 \\
& $\tau_{\mathrm{merge}}$ & 0.82 \\
& $\tau_{\mathrm{new}}$ & $\tau_{\mathrm{merge}}-0.10=0.72$ \\
& $\tau_{\mathrm{novel}}$ & 0.80 \\
& Prototype consolidation threshold $\tau_{\mathrm{proto}}$ & 0.97 \\
& Token-composition temperature $\tau_{\mathrm{temp}}$ & 0.50 \\
& Overlap-ratio threshold & 0.50 \\
& Maximum cross-scale candidates per variable & 16 \\
& Local redundancy window & 3 \\
& EMA momentum $\eta$ & 0.95 \\
\midrule
\multirow{4}{*}{Dictionary}
& Maximum dictionary capacity (non-pretrained) & 128 \\
& Maximum dictionary capacity (pre-training) & 1024 \\
& Minimum novelty support $n_{\min}$ & 6 \\
& Maximum births per maintenance step & 1 \\
\bottomrule
\end{tabular}
\end{table}

\noindent\textbf{Training and Forecasting Details.}
For standard supervised forecasting, we use Adam with an initial learning rate of $1\times10^{-4}$. Following iTransformer~\citep{iTransformer}, we construct training targets by shifting the input window forward by one token. With an input length of $L=672$ and a token length of 96, the input covers time steps $1$-$672$, and the MSE is computed over outputs aligned with time steps $97$-$768$. TFU representations and scene context are constructed from the full observed input window, while token-level attention remains causally masked. The intermediate outputs are therefore supervised on targets within the input window, whereas the final output token predicts the next 96 unobserved time steps. We train for up to 10 epochs using cosine annealing and select the checkpoint with the lowest validation MSE. The early-stopping patience is set to 10 epochs. The batch size is 32 for the four ETT datasets and Weather, 16 for Solar, and 4 for ECL. At inference, only the final output token is used for forecasting. Longer horizons are obtained by appending predicted tokens to the input without accessing future ground-truth observations. The input window is advanced until the requested horizon is covered, and the concatenated predictions are truncated to the required length when necessary. Direct forecasting baselines follow the same rolling forecasting procedure, while autoregressive-token models retain their native decoding interfaces.

\noindent\textbf{Historical-Pattern Retrieval.}
Retrieval is performed independently for each sample within the current input window, without accessing an external database or other samples. Let $\widetilde{X}\in\mathbb{R}^{L\times C}$ denote the input after the model's input normalization. The query contains its last $Q=96$ observations. Candidate segments have length $Q$ and are extracted with a stride of one time step. A candidate is eligible only if its following $P=96$ observations also lie within the current window. We select a shared starting position across variables by minimizing the mean squared distance:
\begin{equation}
\begin{gathered}
s^* =
\underset{1\leq s\leq L-Q-P+1}{\arg\min}
\frac{1}{QC}
\sum_{j=0}^{Q-1}\sum_{c=1}^{C}
\left(
\widetilde{X}_{s+j,c}
-\widetilde{X}_{L-Q+1+j,c}
\right)^2,\\
(Y_{\mathrm{h}})_{r,c}
=
\widetilde{X}_{s^*+Q+r-1,c},
\qquad
1\leq r\leq P,\;
1\leq c\leq C,
\qquad
Y_{\mathrm{h}}\in\mathbb{R}^{P\times C}.
\end{gathered}
\end{equation}
The constraint ensures that the retrieved continuation ends no later than step $L$. The retrieved continuation may overlap the query, but never accesses observations beyond the current window. Retrieval and fusion are performed in the normalized space, followed by the model's inverse normalization. The historical-pattern residual is applied only to the final output token during training and inference. During rolling forecasting, retrieval is repeated on the current window, which may contain previously predicted tokens; future ground-truth observations are never appended.

\noindent\textbf{TFU Learning Configuration.}
Candidate segments are extracted at temporal scales $\mathcal{S}=\{1,2,4\}$, corresponding to spans of 96, 192, and 384 observations. The maximum dictionary capacity is $K=128$ for all non-pretrained experiments. Candidates are extracted with a stride of one token, with at most eight candidates retained per variable at each scale before redundancy suppression. The dictionary maintains an adaptive number of active TFUs, denoted by $K_{\mathrm{active}}\leq K$. New prototypes occupy available slots, while prototype consolidation can reduce the active count; the dictionary is not required to contain $K$ active TFUs. Redundancy suppression uses $\tau_{\mathrm{intra}}=\tau_{\mathrm{cross}}=0.92$ and an overlap-ratio threshold of 0.5, with a local redundancy window of 3 and at most 16 retained cross-scale candidates per variable. For prototype routing, candidates with best-match cosine similarity at least $\tau_{\mathrm{merge}}=0.82$ enter the merge buffer, while those below $\tau_{\mathrm{new}}=0.72$ enter the novelty buffer. The two routing thresholds are coupled by the fixed margin $\tau_{\mathrm{new}}=\tau_{\mathrm{merge}}-0.10$, rather than tuned independently. Candidates with best-match similarity in $[\tau_{\mathrm{new}},\tau_{\mathrm{merge}})$ do not contribute to dictionary updates. Novelty grouping uses $\tau_{\mathrm{novel}}=0.80$. A group must contain at least $n_{\min}=6$ candidates to initialize a new TFU, subject to available dictionary capacity and a limit of one birth per maintenance step. Existing TFUs are refined through grouped EMA updates with momentum $\eta=0.95$. Prototype pairs with cosine similarity at least $\tau_{\mathrm{proto}}=0.97$ are eligible for consolidation to reduce dictionary redundancy. This prototype-to-prototype operation is distinct from candidate routing into the merge buffer. Token-level TFU composition uses a similarity temperature of $\tau_{\mathrm{temp}}=0.5$.

\noindent\textbf{Gradient Propagation.}
Nearest-prototype assignment is discrete, but the selected cosine similarities remain differentiable and determine the token-level composition weights. Forecasting gradients can therefore reach the segment encoder through these weights, with the retained candidate set and prototype indices treated as fixed during backpropagation. This path contributes zero gradient at positions covered by a single candidate or by candidates assigned to the same prototype. Encoder learning through this path thus depends on overlapping candidates with different assigned prototypes. The prototypes are maintained as non-gradient buffers using detached candidate embeddings. No straight-through estimator or additional commitment or reconstruction loss is used.

\noindent\textbf{Dictionary Maintenance and Evaluation.}
During training, retained candidates are matched to active TFUs using cosine similarity. Dictionary maintenance enables grouped EMA updates, novelty birth, and prototype consolidation. Candidates in the merge buffer refine their matched prototypes, supported groups in the novelty buffer can initialize new prototypes, and sufficiently similar prototypes can be consolidated. All dictionary maintenance operations are disabled during validation and testing, so the learned dictionary remains fixed throughout evaluation. In direct source-to-target transfer, the source-trained dictionary is reused without modification by target evaluation samples. During target-domain adaptation, Frozen-TFU keeps both the source dictionary and the backbone fixed and updates only the forecasting head. Updated-TFU keeps the backbone fixed while adapting the forecasting head and enabling all dictionary maintenance operations on the available target-domain training data, including prototype updates, novelty birth, and prototype consolidation. Thus, Updated-TFU adapts both the prototype representations and the dictionary structure, rather than only refining a fixed set of source prototypes. Its comparison with Frozen-TFU evaluates the contribution of adaptive dictionary maintenance under the same target-data budget. The adapted dictionary is subsequently frozen for target-domain evaluation.

\noindent\textbf{Pre-training Configuration.}
We pre-train UniTFU on the combined UTSD and LOTSA corpus after excluding all target evaluation datasets. The backbone architecture, tokenization scheme, temporal scales, and TFU-specific thresholds remain unchanged. The dictionary capacity is increased to $K=1024$ to accommodate the more diverse pre-training corpus. Pre-training uses Adam with an MSE objective, an initial learning rate of $2\times10^{-5}$, an effective global batch size of 4096, and cosine annealing for at most 10 epochs. The resulting checkpoint is used in the pre-trained forecasting evaluation.

\begin{table}[t]
\centering
\caption{Complete direct transfer results under the $672\rightarrow96$ forecasting setting. Each row reports the performance of a model trained on the specified source domain and evaluated on the target domain without adaptation. Average reports the mean across all 20 transfers. \textbf{\textcolor{red}{Red}}: the best, \textbf{\textcolor{blue}{Blue}}: the 2nd best.}
\label{tab:direct_transfer_full}
\begingroup
\scriptsize
\setlength{\tabcolsep}{1.5pt}
\renewcommand{\arraystretch}{1.10}
\newcommand{\directbest}[1]{\textbf{\textcolor{red}{#1}}}
\newcommand{\directsecond}[1]{\textbf{\textcolor{blue}{#1}}}
\resizebox{\textwidth}{!}{%
\begin{tabular}{ll|cc|cc|cc|cc|cc|cc|cc|cc|cc}
% \begin{tabular}{ll*{18}{r}}
\toprule
\multirow{2}{*}{Target} & \multirow{2}{*}{Source} & \multicolumn{2}{c|}{UniTFU} & \multicolumn{2}{c|}{w/ VQ} & \multicolumn{2}{c|}{w/o TFU} & \multicolumn{2}{c|}{GTM} & \multicolumn{2}{c|}{Timer-XL} & \multicolumn{2}{c|}{Timer} & \multicolumn{2}{c|}{UniTime} & \multicolumn{2}{c|}{LangTime} & \multicolumn{2}{c}{LSTPO} \\
\cmidrule(lr){3-20}
& & MSE & MAE & MSE & MAE & MSE & MAE & MSE & MAE & MSE & MAE & MSE & MAE & MSE & MAE & MSE & MAE & MSE & MAE \\
\midrule
\multirow{4}{*}{ETTh1} & ETTm1 & 0.476 & \directbest{0.460} & 0.493 & \directsecond{0.473} & 0.516 & 0.481 & 0.581 & 0.492 & 0.650 & 0.558 & 0.509 & 0.489 & 0.482 & 0.479 & \directbest{0.457} & 0.479 & \directsecond{0.470} & 0.476 \\
 & Weather & \directbest{0.667} & \directbest{0.550} & \directsecond{0.682} & \directsecond{0.554} & 0.695 & 0.566 & 0.707 & 0.563 & 0.723 & 0.583 & 0.698 & 0.563 & 0.718 & 0.567 & 0.697 & 0.567 & 0.731 & \directbest{0.550} \\
 & ECL & \directbest{0.358} & \directbest{0.392} & \directsecond{0.369} & \directsecond{0.396} & \directsecond{0.369} & \directbest{0.392} & 0.482 & 0.475 & 0.402 & 0.418 & 0.378 & 0.399 & 0.420 & 0.430 & 0.418 & 0.408 & 0.403 & 0.414 \\
 & Solar & \directbest{0.645} & \directbest{0.558} & \directsecond{0.699} & \directsecond{0.580} & 0.951 & 0.670 & 0.841 & 0.776 & 0.766 & 0.605 & 0.954 & 0.682 & 0.812 & 0.701 & 0.817 & 0.703 & 0.825 & 0.673 \\
\midrule
\multirow{4}{*}{ETTm1} & ETTh1 & \directbest{0.636} & \directbest{0.530} & \directsecond{0.661} & \directsecond{0.543} & 0.783 & 0.556 & 0.857 & 0.595 & 0.748 & 0.562 & 0.794 & 0.568 & 0.755 & 0.574 & 0.727 & 0.549 & 0.757 & 0.567 \\
 & Weather & 1.268 & 0.702 & 1.264 & 0.701 & 0.940 & 0.608 & 0.803 & 0.554 & 0.858 & 0.605 & 0.935 & 0.610 & 0.755 & 0.554 & \directsecond{0.749} & \directsecond{0.551} & \directbest{0.736} & \directbest{0.531} \\
 & ECL & \directbest{0.656} & \directbest{0.535} & 0.746 & 0.555 & 0.890 & 0.608 & 0.971 & 0.680 & 0.868 & 0.594 & 0.902 & 0.613 & 0.766 & 0.567 & 0.767 & 0.567 & \directsecond{0.733} & \directsecond{0.538} \\
 & Solar & \directbest{0.956} & \directbest{0.638} & \directsecond{1.003} & \directsecond{0.668} & 1.776 & 0.873 & 1.155 & 0.750 & 1.103 & 0.695 & 1.786 & 0.885 & 1.224 & 0.788 & 1.182 & 0.747 & 1.193 & 0.772 \\
\midrule
\multirow{4}{*}{Weather} & ETTh1 & 0.215 & 0.273 & 0.221 & 0.279 & \directbest{0.198} & \directbest{0.255} & 0.314 & 0.347 & 0.223 & 0.284 & 0.209 & \directsecond{0.269} & 0.213 & 0.282 & 0.210 & 0.279 & \directsecond{0.207} & 0.278 \\
 & ETTm1 & 0.224 & 0.276 & 0.227 & 0.278 & 0.222 & 0.273 & 0.307 & 0.336 & 0.240 & 0.296 & 0.230 & 0.283 & 0.208 & \directsecond{0.270} & \directbest{0.199} & \directbest{0.266} & \directsecond{0.204} & \directbest{0.266} \\
 & ECL & \directbest{0.215} & \directbest{0.270} & \directsecond{0.223} & \directsecond{0.272} & 0.281 & 0.318 & 0.347 & 0.367 & 0.270 & 0.319 & 0.232 & 0.274 & 0.268 & 0.318 & 0.270 & 0.317 & 0.259 & 0.323 \\
 & Solar & \directbest{0.324} & \directbest{0.330} & 0.330 & \directsecond{0.337} & 0.368 & 0.361 & 0.335 & 0.353 & 0.375 & 0.354 & 0.404 & 0.374 & 0.330 & 0.349 & 0.331 & 0.342 & \directsecond{0.328} & 0.342 \\
\midrule
\multirow{4}{*}{ECL} & ETTh1 & \directbest{0.211} & \directbest{0.302} & \directsecond{0.214} & \directsecond{0.303} & 0.220 & 0.306 & 0.348 & 0.426 & 0.259 & 0.353 & 0.217 & 0.305 & 0.393 & 0.471 & 0.388 & 0.470 & 0.382 & 0.458 \\
 & ETTm1 & 0.646 & 0.584 & 0.661 & 0.601 & \directbest{0.385} & \directbest{0.465} & 0.664 & 0.608 & 0.620 & 0.602 & \directsecond{0.397} & \directsecond{0.475} & 0.533 & 0.554 & 0.531 & 0.535 & 0.527 & 0.533 \\
 & Weather & \directsecond{0.688} & \directsecond{0.663} & 0.697 & 0.670 & 0.825 & 0.750 & \directbest{0.583} & \directbest{0.566} & 0.807 & 0.734 & 0.836 & 0.755 & 1.053 & 0.804 & 1.029 & 0.773 & 1.040 & 0.794 \\
 & Solar & \directsecond{0.772} & 0.692 & 0.832 & 0.704 & 0.990 & 0.780 & 1.013 & 0.769 & \directbest{0.763} & \directbest{0.680} & 0.999 & 0.786 & 0.795 & 0.711 & 0.797 & 0.714 & 0.778 & \directsecond{0.689} \\
\midrule
\multirow{4}{*}{Solar} & ETTh1 & \directbest{0.790} & \directbest{0.665} & \directsecond{0.808} & \directsecond{0.674} & 0.860 & 0.682 & 0.841 & 0.686 & 0.878 & 0.707 & 0.853 & 0.697 & 0.822 & 0.750 & 0.810 & 0.745 & 0.809 & 0.767 \\
 & ETTm1 & 0.742 & 0.627 & 0.748 & 0.639 & \directbest{0.600} & 0.564 & 0.669 & \directbest{0.536} & 0.966 & 0.727 & \directsecond{0.615} & \directsecond{0.557} & 1.074 & 0.837 & 1.041 & 0.836 & 1.046 & 0.817 \\
 & Weather & \directbest{0.396} & \directbest{0.452} & \directsecond{0.403} & \directsecond{0.453} & 0.408 & 0.461 & 0.459 & 0.458 & 0.495 & 0.503 & 0.419 & 0.456 & 0.881 & 0.774 & 0.873 & 0.735 & 0.839 & 0.759 \\
 & ECL & \directbest{0.551} & \directbest{0.505} & \directsecond{0.669} & \directsecond{0.594} & 0.932 & 0.691 & 1.434 & 0.932 & 1.108 & 0.827 & 0.938 & 0.697 & 0.763 & 0.649 & 0.742 & 0.620 & 0.750 & 0.649 \\
\midrule
\multicolumn{2}{l|}{Average} & \directbest{0.572} & \directbest{0.500} & \directsecond{0.598} & \directsecond{0.514} & 0.660 & 0.533 & 0.686 & 0.563 & 0.656 & 0.550 & 0.665 & 0.537 & 0.663 & 0.571 & 0.652 & 0.560 & 0.651 & 0.560 \\
\bottomrule
\end{tabular}%
}
\endgroup
\end{table}

\section{Full Cross-Domain Forecasting Results}
\label{App_cross_domain}

\subsection{Direct Transfer} \label{App_direct_transfer}

Table~\ref{tab:direct_transfer_full} presents the complete direct transfer results across ETTh1, ETTm1, Weather, ECL, and Solar under the $672\rightarrow96$ forecasting setting. For each target domain, we report the results of models trained separately on each of the remaining four source domains. All model parameters and, when applicable, dictionary or codebook states remain fixed during target-domain evaluation. Implementation details are provided in
Appendix~\ref{App_implementation}. Averaging the results over the four source domains for each target domain gives the aggregated results in
Table~\ref{zero_transfer}. With UniTime, LangTime, and LSTPO included, UniTFU achieves the lowest target-averaged MSE on four of five domains and the lowest MAE on all five, with the best overall averages and the best MSE and MAE simultaneously in 12 of 20 transfers. Compared with w/o TFU and w/ VQ, UniTFU reduces overall average MSE by 13.4\% and 4.3\%, respectively, supporting the contribution of TFU modeling and the adaptive dictionary design to direct cross-domain transfer. Performance nevertheless varies with the source domain. For example, UniTFU underperforms the baselines when transferring from Weather to ETTm1, indicating that its overall advantage does not extend to every transfer setting.

\subsection{Target Domain Adaptation} \label{App_target_adaptation}

Table~\ref{tab:target_adaptation_full} presents the complete
target-domain adaptation results under the $672\rightarrow96$
forecasting setting. For each target domain, models trained separately on the remaining four source domains are adapted using 10\% of the
target-domain training samples. We report standard few-shot results for Timer, Timer-XL, GTM, UniTime, LangTime, LSTPO, and UniTFU, together with Frozen-TFU and Updated-TFU.
Frozen-TFU updates only the forecasting head, whereas Updated-TFU additionally updates the TFU dictionary; both keep the remaining backbone fixed. All model parameters and TFU prototypes remain fixed during subsequent target-domain evaluation. Implementation details are provided in
Appendix~\ref{App_implementation}. Averaging the results over the four source domains for each target domain gives the aggregated results in Table~\ref{few_shot_transfer}. Updated-TFU reduces MSE in 18 of the 20 adaptation settings and MAE in 19 compared with Frozen-TFU, yielding overall reductions of 3.7\% and 3.9\%, respectively. These results support the refinement of reusable temporal structures with limited target-domain data. On ETTm1, Updated-TFU outperforms standard few-shot UniTFU on both metrics for every source domain. On ECL and Solar, however, it improves over Frozen-TFU but remains behind standard few-shot UniTFU and Timer-XL for every source domain. These results indicate that the benefits of TFU dictionary adaptation vary across target domains and that some targets require broader model adaptation.

\begin{table}[t]
\centering
\caption{Complete target-domain adaptation results using 10\% of the target-domain training samples under the $672\rightarrow96$ forecasting setting. Frozen-TFU and Updated-TFU are defined in the text. Average reports the mean across all 20 adaptation settings. \textbf{\textcolor{red}{Red}}: the best, \textbf{\textcolor{blue}{Blue}}: the 2nd best.}
\label{tab:target_adaptation_full}
\begingroup
\scriptsize
\setlength{\tabcolsep}{1.5pt}
\renewcommand{\arraystretch}{1.10}
\newcommand{\adaptbest}[1]{\textbf{\textcolor{red}{#1}}}
\newcommand{\adaptsecond}[1]{\textbf{\textcolor{blue}{#1}}}
\resizebox{\textwidth}{!}{%
\begin{tabular}{ll*{18}{r}}
% \begin{tabular}{|l|l|*{18}{r|}}
\toprule
& & \multicolumn{6}{c|}{10\% Few-shot} & \multicolumn{2}{c|}{Frozen-TFU} & \multicolumn{2}{c|}{Updated-TFU} & \multicolumn{8}{c}{10\% Few-shot} \\
\cmidrule(lr){3-8}\cmidrule(lr){9-10}\cmidrule(lr){11-12}\cmidrule(lr){13-20}
\multirow{2}{*}{Target} & \multirow{2}{*}{Source} & \multicolumn{2}{c|}{Timer} & \multicolumn{2}{c|}{Timer-XL} & \multicolumn{2}{c|}{GTM} & \multicolumn{2}{c|}{UniTFU} & \multicolumn{2}{c|}{UniTFU} & \multicolumn{2}{c|}{UniTFU} & \multicolumn{2}{c|}{UniTime} & \multicolumn{2}{c|}{LangTime} & \multicolumn{2}{c}{LSTPO} \\
\cmidrule(lr){3-20}
& & MSE & MAE & MSE & MAE & MSE & MAE & MSE & MAE & MSE & MAE & MSE & MAE & MSE & MAE & MSE & MAE & MSE & MAE \\
\midrule
\multirow{4}{*}{ETTh1} & ETTm1 & 0.446 & 0.456 & 0.482 & 0.481 & 0.518 & 0.474 & 0.447 & \adaptsecond{0.448} & 0.445 & \adaptbest{0.445} & 0.458 & 0.463 & \adaptsecond{0.437} & 0.454 & \adaptbest{0.432} & 0.463 & 0.446 & 0.473 \\
 & Weather & 0.594 & 0.535 & 0.581 & 0.528 & \adaptsecond{0.522} & 0.503 & 0.566 & 0.513 & 0.557 & 0.502 & \adaptbest{0.490} & \adaptbest{0.488} & 0.525 & \adaptsecond{0.496} & 0.523 & 0.509 & 0.555 & 0.515 \\
 & ECL & 0.378 & 0.410 & 0.391 & 0.413 & 0.418 & 0.430 & \adaptsecond{0.355} & \adaptsecond{0.392} & 0.362 & 0.399 & \adaptbest{0.340} & \adaptbest{0.370} & 0.420 & 0.445 & 0.408 & 0.420 & 0.436 & 0.471 \\
 & Solar & 0.616 & 0.537 & \adaptsecond{0.510} & \adaptbest{0.487} & 0.644 & 0.552 & 0.559 & 0.513 & 0.546 & 0.500 & \adaptbest{0.499} & \adaptsecond{0.491} & 0.533 & 0.498 & 0.517 & 0.523 & 0.550 & 0.495 \\
\midrule
\multirow{4}{*}{ETTm1} & ETTh1 & 0.557 & 0.542 & 0.628 & 0.500 & 0.704 & 0.500 & \adaptsecond{0.426} & 0.423 & \adaptbest{0.421} & \adaptsecond{0.422} & 0.523 & 0.464 & 0.443 & \adaptbest{0.419} & 0.589 & 0.453 & 0.440 & 0.439 \\
 & Weather & 0.556 & 0.478 & 0.493 & 0.460 & 0.578 & 0.478 & \adaptsecond{0.446} & \adaptsecond{0.435} & \adaptbest{0.433} & \adaptbest{0.426} & 0.606 & 0.481 & 0.490 & 0.453 & 0.491 & 0.438 & 0.498 & 0.458 \\
 & ECL & 0.651 & 0.537 & 0.552 & 0.473 & 0.752 & 0.529 & \adaptbest{0.436} & 0.428 & 0.440 & \adaptsecond{0.424} & 0.547 & 0.477 & 0.488 & 0.462 & 0.502 & 0.481 & \adaptsecond{0.439} & \adaptbest{0.399} \\
 & Solar & 0.534 & 0.470 & 0.605 & 0.497 & \adaptsecond{0.465} & \adaptbest{0.441} & 0.478 & 0.448 & \adaptbest{0.464} & \adaptsecond{0.444} & 0.543 & 0.472 & 0.522 & 0.466 & 0.498 & 0.452 & 0.499 & 0.452 \\
\midrule
\multirow{4}{*}{Weather} & ETTh1 & 0.176 & \adaptbest{0.227} & 0.177 & 0.233 & 0.179 & 0.231 & 0.188 & 0.247 & 0.184 & 0.243 & \adaptbest{0.174} & \adaptsecond{0.230} & 0.186 & 0.244 & 0.184 & 0.240 & \adaptsecond{0.175} & 0.246 \\
 & ETTm1 & \adaptbest{0.171} & 0.219 & 0.178 & 0.230 & \adaptbest{0.171} & 0.222 & 0.184 & 0.242 & 0.181 & 0.239 & \adaptsecond{0.172} & \adaptsecond{0.215} & 0.180 & 0.222 & 0.176 & \adaptbest{0.213} & 0.190 & 0.225 \\
 & ECL & \adaptbest{0.172} & \adaptbest{0.216} & \adaptbest{0.172} & \adaptsecond{0.220} & 0.187 & 0.235 & 0.192 & 0.248 & 0.185 & 0.242 & \adaptsecond{0.183} & \adaptsecond{0.220} & 0.188 & 0.223 & 0.189 & 0.236 & 0.192 & 0.221 \\
 & Solar & 0.183 & 0.233 & \adaptbest{0.178} & 0.229 & \adaptbest{0.178} & 0.232 & 0.197 & 0.254 & 0.194 & 0.250 & 0.180 & \adaptsecond{0.226} & 0.185 & 0.235 & \adaptsecond{0.179} & \adaptbest{0.223} & 0.181 & 0.249 \\
\midrule
\multirow{4}{*}{ECL} & ETTh1 & 0.173 & 0.282 & 0.171 & \adaptsecond{0.264} & 0.183 & 0.280 & 0.217 & 0.315 & 0.211 & 0.308 & \adaptsecond{0.165} & \adaptbest{0.259} & 0.174 & 0.269 & \adaptbest{0.164} & 0.272 & 0.181 & 0.283 \\
 & ETTm1 & 0.167 & 0.267 & 0.190 & 0.283 & 0.188 & 0.285 & 0.232 & 0.332 & 0.215 & 0.315 & 0.164 & \adaptbest{0.259} & \adaptsecond{0.160} & 0.270 & 0.163 & \adaptsecond{0.262} & \adaptbest{0.155} & 0.279 \\
 & Weather & \adaptbest{0.154} & \adaptsecond{0.253} & \adaptbest{0.154} & \adaptbest{0.250} & 0.187 & 0.284 & 0.233 & 0.333 & 0.217 & 0.317 & \adaptsecond{0.162} & 0.258 & 0.167 & 0.280 & 0.168 & 0.279 & 0.171 & 0.285 \\
 & Solar & \adaptsecond{0.168} & 0.267 & 0.170 & 0.265 & 0.195 & 0.292 & 0.256 & 0.354 & 0.223 & 0.322 & \adaptbest{0.163} & \adaptbest{0.259} & 0.186 & 0.276 & 0.190 & \adaptsecond{0.260} & 0.197 & 0.277 \\
\midrule
\multirow{4}{*}{Solar} & ETTh1 & \adaptsecond{0.199} & 0.267 & \adaptbest{0.180} & \adaptbest{0.237} & 0.209 & \adaptsecond{0.241} & 0.266 & 0.346 & 0.240 & 0.315 & 0.202 & 0.249 & 0.205 & 0.292 & 0.206 & 0.280 & 0.208 & 0.304 \\
 & ETTm1 & \adaptbest{0.192} & 0.256 & \adaptbest{0.192} & 0.252 & 0.203 & \adaptsecond{0.237} & 0.250 & 0.326 & 0.228 & 0.292 & \adaptsecond{0.201} & \adaptbest{0.234} & 0.209 & 0.269 & 0.212 & 0.257 & 0.220 & 0.274 \\
 & Weather & \adaptbest{0.187} & 0.248 & 0.191 & \adaptbest{0.230} & \adaptsecond{0.189} & \adaptbest{0.230} & 0.241 & 0.307 & 0.223 & 0.274 & 0.216 & \adaptsecond{0.244} & 0.211 & 0.277 & 0.208 & 0.266 & 0.210 & 0.288 \\
 & ECL & 0.194 & 0.246 & \adaptbest{0.188} & \adaptsecond{0.240} & 0.213 & 0.245 & 0.280 & 0.338 & 0.240 & 0.283 & \adaptsecond{0.190} & \adaptbest{0.235} & 0.257 & 0.261 & 0.259 & 0.253 & 0.272 & 0.268 \\
\midrule
\multicolumn{2}{l}{Average} & 0.323 & 0.347 & 0.319 & \adaptsecond{0.339} & 0.344 & 0.346 & 0.322 & 0.362 & \adaptsecond{0.310} & 0.348 & \adaptbest{0.309} & \adaptbest{0.330} & \adaptbest{0.309} & 0.341 & 0.313 & \adaptsecond{0.339} & 0.311 & 0.345 \\
\bottomrule
\end{tabular}%
}
\endgroup
\end{table}

\begin{table}[t]
\centering
\caption{Complete multivariate rolling forecasting results with
a lookback length of 672 and prediction lengths $\{96,192,336,720\}$. \textbf{\textcolor{red}{Red}}: the best, \textbf{\textcolor{blue}{Blue}}: the 2nd best.}
\label{tab:multivariate_full}
\begingroup
\newcommand{\mvbest}[1]{\textbf{\textcolor{red}{#1}}}
\newcommand{\mvsecond}[1]{\textbf{\textcolor{blue}{#1}}}
\setlength{\tabcolsep}{2pt}
\renewcommand{\arraystretch}{1.2}
\resizebox{\textwidth}{!}{%
\begin{tabular}{cc|rr|rr|rr|rr|rr|rr|rr|rr}
\toprule
\multicolumn{2}{c|}{Type}
& \multicolumn{8}{c|}{TSFM}
& \multicolumn{8}{c}{Forecasting Model} \\
\midrule
\multicolumn{2}{c|}{Models}
& \multicolumn{2}{c|}{\textbf{UniTFU}}
& \multicolumn{2}{c|}{GTM}
& \multicolumn{2}{c|}{Timer-XL}
& \multicolumn{2}{c|}{Timer}
& \multicolumn{2}{c|}{PhaseFormer}
& \multicolumn{2}{c|}{iTransformer}
& \multicolumn{2}{c|}{DLinear}
& \multicolumn{2}{c}{PatchTST} \\
\midrule
Dataset & $H$
& MSE & MAE & MSE & MAE & MSE & MAE & MSE & MAE
& MSE & MAE & MSE & MAE & MSE & MAE & MSE & MAE \\
\midrule
\multirow{5}{*}{\rotatebox[origin=c]{90}{ETTh1}} & 96 & \mvbest{0.360} & \mvbest{0.396} & \mvsecond{0.362} & \mvsecond{0.397} & 0.365 & 0.399 & 0.371 & 0.402 & 0.374 & \mvsecond{0.397} & 0.382 & 0.413 & 0.373 & 0.402 & 0.376 & 0.405 \\
 & 192 & \mvsecond{0.392} & \mvbest{0.418} & \mvbest{0.389} & \mvsecond{0.419} & 0.407 & 0.427 & 0.410 & 0.427 & 0.411 & 0.420 & 0.413 & 0.432 & 0.411 & 0.427 & 0.407 & 0.427 \\
 & 336 & \mvsecond{0.410} & \mvbest{0.430} & \mvbest{0.406} & \mvsecond{0.434} & 0.429 & 0.443 & 0.434 & 0.443 & 0.435 & 0.435 & 0.432 & 0.445 & 0.446 & 0.454 & 0.420 & 0.438 \\
 & 720 & \mvbest{0.423} & \mvbest{0.446} & 0.433 & 0.465 & 0.436 & 0.461 & 0.458 & 0.459 & 0.451 & \mvsecond{0.456} & 0.447 & 0.470 & 0.515 & 0.522 & \mvsecond{0.424} & \mvsecond{0.456} \\
\cmidrule(lr){2-18}
 & Avg. & \mvbest{0.396} & \mvbest{0.423} & \mvsecond{0.398} & 0.429 & 0.409 & 0.433 & 0.418 & 0.433 & 0.418 & \mvsecond{0.427} & 0.419 & 0.440 & 0.436 & 0.451 & 0.407 & 0.432 \\
\midrule
\multirow{5}{*}{\rotatebox[origin=c]{90}{ETTh2}} & 96 & 0.297 & \mvsecond{0.350} & \mvsecond{0.292} & \mvsecond{0.350} & 0.299 & 0.353 & \mvbest{0.286} & \mvbest{0.345} & 0.323 & 0.381 & 0.329 & 0.377 & 0.308 & 0.373 & \mvsecond{0.292} & 0.351 \\
 & 192 & \mvbest{0.360} & \mvbest{0.394} & 0.368 & 0.402 & 0.367 & \mvsecond{0.400} & \mvsecond{0.364} & \mvsecond{0.400} & 0.367 & 0.408 & 0.391 & 0.418 & 0.414 & 0.441 & 0.371 & \mvsecond{0.400} \\
 & 336 & \mvsecond{0.384} & \mvbest{0.420} & 0.400 & 0.432 & 0.393 & 0.427 & 0.409 & 0.440 & \mvbest{0.382} & \mvsecond{0.422} & 0.412 & 0.436 & 0.528 & 0.508 & 0.405 & 0.432 \\
 & 720 & 0.422 & 0.454 & 0.438 & 0.465 & 0.444 & 0.470 & 0.455 & 0.482 & \mvbest{0.403} & \mvbest{0.442} & 0.448 & 0.463 & 0.830 & 0.653 & \mvsecond{0.419} & \mvsecond{0.453} \\
\cmidrule(lr){2-18}
 & Avg. & \mvbest{0.366} & \mvbest{0.405} & 0.375 & 0.412 & 0.376 & 0.413 & 0.379 & 0.417 & \mvsecond{0.369} & 0.413 & 0.395 & 0.424 & 0.520 & 0.494 & 0.372 & \mvsecond{0.409} \\
\midrule
\multirow{5}{*}{\rotatebox[origin=c]{90}{ETTm1}} & 96 & \mvbest{0.272} & \mvbest{0.327} & 0.285 & \mvsecond{0.334} & 0.296 & 0.351 & \mvsecond{0.282} & 0.338 & 0.305 & 0.349 & 0.325 & 0.373 & 0.306 & 0.349 & 0.299 & 0.357 \\
 & 192 & \mvbest{0.315} & \mvbest{0.358} & \mvsecond{0.326} & \mvsecond{0.360} & 0.346 & 0.381 & 0.327 & 0.366 & 0.337 & 0.370 & 0.365 & 0.397 & 0.336 & 0.367 & 0.340 & 0.381 \\
 & 336 & \mvbest{0.350} & \mvbest{0.382} & 0.361 & \mvsecond{0.383} & 0.385 & 0.404 & \mvsecond{0.360} & 0.387 & 0.367 & 0.390 & 0.397 & 0.416 & 0.365 & 0.385 & 0.376 & 0.402 \\
 & 720 & \mvsecond{0.412} & \mvsecond{0.418} & \mvbest{0.411} & \mvsecond{0.418} & 0.449 & 0.441 & 0.418 & 0.420 & 0.429 & 0.424 & 0.460 & 0.449 & 0.417 & \mvbest{0.416} & 0.438 & 0.437 \\
\cmidrule(lr){2-18}
 & Avg. & \mvbest{0.337} & \mvbest{0.371} & \mvsecond{0.346} & \mvsecond{0.374} & 0.369 & 0.394 & 0.347 & 0.378 & 0.360 & 0.383 & 0.387 & 0.409 & 0.356 & 0.379 & 0.363 & 0.394 \\
\midrule
\multirow{5}{*}{\rotatebox[origin=c]{90}{ETTm2}} & 96 & 0.182 & 0.265 & \mvsecond{0.168} & \mvsecond{0.256} & 0.184 & 0.270 & \mvbest{0.167} & \mvbest{0.255} & 0.177 & 0.265 & 0.189 & 0.276 & \mvbest{0.167} & 0.262 & 0.173 & 0.265 \\
 & 192 & 0.246 & 0.310 & 0.230 & \mvsecond{0.300} & 0.242 & 0.309 & \mvbest{0.226} & \mvbest{0.295} & 0.238 & 0.304 & 0.259 & 0.320 & 0.230 & 0.312 & \mvsecond{0.229} & 0.301 \\
 & 336 & 0.304 & 0.351 & 0.287 & 0.340 & 0.297 & 0.343 & \mvbest{0.279} & \mvbest{0.332} & 0.291 & 0.339 & 0.321 & 0.360 & 0.300 & 0.362 & \mvsecond{0.280} & \mvsecond{0.334} \\
 & 720 & 0.408 & 0.417 & 0.378 & 0.398 & 0.380 & 0.396 & \mvsecond{0.369} & \mvsecond{0.391} & 0.376 & 0.394 & 0.415 & 0.418 & 0.437 & 0.449 & \mvbest{0.368} & \mvbest{0.390} \\
\cmidrule(lr){2-18}
 & Avg. & 0.285 & 0.336 & 0.266 & 0.324 & 0.276 & 0.330 & \mvbest{0.260} & \mvbest{0.318} & 0.271 & 0.326 & 0.296 & 0.344 & 0.284 & 0.346 & \mvsecond{0.263} & \mvsecond{0.323} \\
\midrule
\multirow{5}{*}{\rotatebox[origin=c]{90}{ECL}} & 96 & \mvbest{0.132} & \mvbest{0.229} & 0.141 & 0.241 & 0.135 & \mvsecond{0.230} & 0.137 & 0.232 & 0.138 & 0.232 & \mvsecond{0.133} & \mvsecond{0.230} & 0.135 & 0.233 & 0.134 & 0.234 \\
 & 192 & \mvbest{0.150} & \mvbest{0.245} & 0.160 & 0.259 & 0.153 & 0.247 & 0.153 & \mvsecond{0.246} & \mvsecond{0.152} & \mvbest{0.245} & \mvsecond{0.152} & \mvbest{0.245} & 0.154 & 0.247 & 0.153 & 0.253 \\
 & 336 & \mvbest{0.166} & \mvsecond{0.263} & 0.178 & 0.278 & 0.173 & 0.267 & 0.171 & 0.265 & 0.168 & \mvbest{0.261} & 0.169 & \mvsecond{0.263} & \mvsecond{0.167} & 0.264 & 0.175 & 0.276 \\
 & 720 & \mvbest{0.207} & \mvsecond{0.299} & 0.214 & 0.308 & 0.215 & 0.304 & 0.214 & 0.302 & 0.212 & 0.303 & \mvsecond{0.208} & \mvsecond{0.299} & 0.209 & \mvbest{0.298} & 0.229 & 0.325 \\
\cmidrule(lr){2-18}
 & Avg. & \mvbest{0.164} & \mvbest{0.259} & 0.173 & 0.272 & 0.169 & 0.262 & 0.169 & 0.261 & 0.168 & \mvsecond{0.260} & \mvsecond{0.166} & \mvbest{0.259} & \mvsecond{0.166} & 0.261 & 0.173 & 0.272 \\
\midrule
\multirow{5}{*}{\rotatebox[origin=c]{90}{Weather}} & 96 & 0.158 & \mvsecond{0.206} & 0.161 & 0.212 & 0.162 & 0.213 & \mvsecond{0.155} & \mvbest{0.204} & 0.159 & 0.216 & 0.162 & 0.216 & 0.169 & 0.228 & \mvbest{0.154} & 0.207 \\
 & 192 & \mvsecond{0.203} & \mvbest{0.248} & 0.208 & 0.253 & 0.209 & 0.257 & 0.204 & 0.250 & \mvsecond{0.203} & 0.256 & 0.208 & 0.257 & 0.212 & 0.268 & \mvbest{0.200} & \mvsecond{0.249} \\
 & 336 & 0.256 & \mvbest{0.288} & 0.261 & 0.292 & 0.263 & 0.298 & 0.264 & 0.296 & \mvsecond{0.253} & 0.295 & 0.259 & 0.296 & 0.257 & 0.307 & \mvbest{0.251} & \mvsecond{0.289} \\
 & 720 & \mvsecond{0.324} & \mvbest{0.336} & 0.330 & \mvsecond{0.340} & 0.333 & 0.346 & 0.349 & 0.352 & 0.325 & 0.345 & 0.327 & 0.342 & \mvbest{0.320} & 0.361 & \mvbest{0.320} & \mvsecond{0.340} \\
\cmidrule(lr){2-18}
 & Avg. & \mvsecond{0.235} & \mvbest{0.270} & 0.240 & 0.274 & 0.242 & 0.279 & 0.243 & 0.276 & \mvsecond{0.235} & 0.278 & 0.239 & 0.278 & 0.240 & 0.291 & \mvbest{0.231} & \mvsecond{0.271} \\
\midrule
\multirow{5}{*}{\rotatebox[origin=c]{90}{Solar}} & 96 & \mvbest{0.182} & \mvbest{0.220} & 0.193 & \mvsecond{0.221} & 0.186 & 0.238 & 0.192 & 0.243 & 0.193 & 0.260 & 0.192 & 0.258 & 0.192 & 0.257 & \mvsecond{0.185} & 0.278 \\
 & 192 & \mvsecond{0.202} & \mvsecond{0.246} & 0.220 & \mvbest{0.239} & 0.213 & 0.258 & 0.215 & 0.261 & 0.211 & 0.270 & 0.217 & 0.282 & 0.213 & 0.274 & \mvbest{0.201} & 0.288 \\
 & 336 & \mvbest{0.215} & \mvbest{0.257} & 0.247 & \mvbest{0.257} & 0.231 & \mvsecond{0.271} & 0.231 & 0.274 & 0.226 & 0.279 & 0.226 & 0.300 & 0.233 & 0.290 & \mvsecond{0.217} & 0.298 \\
 & 720 & \mvbest{0.235} & \mvbest{0.276} & 0.262 & 0.283 & 0.244 & \mvsecond{0.281} & 0.241 & 0.285 & \mvbest{0.235} & 0.282 & 0.240 & 0.320 & 0.245 & 0.306 & \mvsecond{0.238} & 0.311 \\
\cmidrule(lr){2-18}
 & Avg. & \mvbest{0.209} & \mvbest{0.250} & 0.231 & \mvbest{0.250} & 0.219 & \mvsecond{0.262} & 0.220 & 0.266 & 0.216 & 0.273 & 0.219 & 0.290 & 0.221 & 0.282 & \mvsecond{0.210} & 0.294 \\
\midrule
\multicolumn{2}{c|}{1st Count} & 13 & 17 & 3 & 2 & 0 & 0 & 4 & 5 & 3 & 3 & 0 & 1 & 2 & 2 & 6 & 1 \\
\bottomrule
\end{tabular}%
}
\endgroup
\end{table}

\section{Full Multivariate Time Series Forecasting Results}
\label{App_multivariate}

Table~\ref{tab:multivariate_full} presents the complete results under standard supervised multivariate rolling forecasting. The lookback length is 672, and the prediction lengths are $\{96,192,336,720\}$. For each dataset, we report MSE and MAE at each prediction length and their averages over the four prediction lengths. Implementation details are provided in Appendix~\ref{App_implementation}. UniTFU achieves the lowest average MSE and MAE on ETTh1, ETTh2, ETTm1, ECL, and Solar, and the lowest average MAE on Weather. It obtains the best MSE in 13 of the 28 forecasting settings and the best MAE in 17. On ECL, UniTFU achieves the lowest MSE at all four prediction lengths, while on Solar it achieves the lowest MSE at three. However, it remains behind several baselines on ETTm2, indicating that its improvements are not uniform across datasets.

\begin{table}[t]
\centering
\caption{Complete pretrained forecasting results with a lookback
length of 672 and prediction horizons $\{96,192,336,720\}$. A dash indicates exclusion because the dataset is included in the corresponding model's pre-training data. \textbf{\textcolor{red}{Red}}: the best, \textbf{\textcolor{blue}{Blue}}: the 2nd best.}
\label{tab:pretrained_full}
\begingroup
\setlength{\tabcolsep}{2pt}
\renewcommand{\arraystretch}{1.08}
\resizebox{\textwidth}{!}{%
\begin{tabular}{cc|rr|rr|rr|rr|rr|rr|rr|rr}
\toprule
\multicolumn{2}{c|}{Models}
& \multicolumn{2}{c|}{\textbf{UniTFU}}
& \multicolumn{2}{c|}{Aurora}
& \multicolumn{2}{c|}{Sundial}
& \multicolumn{2}{c|}{TiRex}
& \multicolumn{2}{c|}{Time-MoE}
& \multicolumn{2}{c|}{MOIRAI}
& \multicolumn{2}{c|}{TimesFM}
& \multicolumn{2}{c}{Chronos} \\
\midrule
Dataset & $H$ & MSE & MAE & MSE & MAE & MSE & MAE & MSE & MAE & MSE & MAE & MSE & MAE & MSE & MAE & MSE & MAE \\
\midrule
\multirow{5}{*}{\rotatebox[origin=c]{90}{ETTh1}} & 96 & \textbf{\textcolor{red}{0.341}} & \textbf{\textcolor{red}{0.376}} & \textbf{\textcolor{blue}{0.342}} & 0.386 & 0.343 & 0.379 & 0.379 & \textbf{\textcolor{blue}{0.377}} & 0.386 & 0.398 & 0.394 & 0.396 & 0.373 & 0.381 & 0.513 & 0.428 \\
 & 192 & \textbf{\textcolor{red}{0.374}} & \textbf{\textcolor{red}{0.398}} & 0.378 & 0.411 & \textbf{\textcolor{blue}{0.376}} & \textbf{\textcolor{blue}{0.400}} & 0.428 & 0.406 & 0.441 & 0.420 & 0.428 & 0.418 & 0.414 & 0.405 & 0.578 & 0.464 \\
 & 336 & \textbf{\textcolor{red}{0.389}} & \textbf{\textcolor{red}{0.408}} & 0.401 & 0.426 & \textbf{\textcolor{blue}{0.391}} & \textbf{\textcolor{blue}{0.410}} & 0.463 & 0.424 & 0.476 & 0.442 & 0.447 & 0.428 & 0.438 & 0.418 & 0.612 & 0.488 \\
 & 720 & \textbf{\textcolor{red}{0.416}} & \textbf{\textcolor{blue}{0.431}} & 0.425 & 0.449 & \textbf{\textcolor{blue}{0.418}} & 0.433 & 0.479 & 0.460 & 0.498 & 0.474 & 0.534 & 0.523 & 0.425 & \textbf{\textcolor{red}{0.429}} & 0.611 & 0.510 \\
\cmidrule(lr){2-18}
 & Avg. & \textbf{\textcolor{red}{0.380}} & \textbf{\textcolor{red}{0.403}} & 0.387 & 0.418 & \textbf{\textcolor{blue}{0.382}} & \textbf{\textcolor{blue}{0.406}} & 0.437 & 0.417 & 0.450 & 0.434 & 0.451 & 0.441 & 0.413 & 0.408 & 0.579 & 0.473 \\
\midrule
\multirow{5}{*}{\rotatebox[origin=c]{90}{ETTh2}} & 96 & 0.299 & 0.337 & \textbf{\textcolor{red}{0.273}} & \textbf{\textcolor{blue}{0.334}} & \textbf{\textcolor{blue}{0.282}} & \textbf{\textcolor{blue}{0.334}} & 0.296 & \textbf{\textcolor{red}{0.326}} & 0.308 & 0.350 & 0.316 & 0.347 & 0.304 & 0.337 & 0.354 & 0.373 \\
 & 192 & 0.355 & 0.378 & \textbf{\textcolor{red}{0.339}} & 0.377 & \textbf{\textcolor{blue}{0.343}} & \textbf{\textcolor{blue}{0.376}} & 0.353 & \textbf{\textcolor{red}{0.368}} & 0.364 & 0.394 & 0.383 & 0.392 & 0.371 & 0.383 & 0.429 & 0.418 \\
 & 336 & 0.371 & 0.400 & 0.376 & 0.408 & \textbf{\textcolor{blue}{0.369}} & \textbf{\textcolor{blue}{0.398}} & \textbf{\textcolor{red}{0.354}} & \textbf{\textcolor{red}{0.379}} & 0.388 & 0.401 & 0.412 & 0.416 & 0.400 & 0.409 & 0.476 & 0.448 \\
 & 720 & \textbf{\textcolor{red}{0.368}} & \textbf{\textcolor{red}{0.401}} & 0.401 & 0.436 & 0.390 & 0.422 & \textbf{\textcolor{blue}{0.370}} & \textbf{\textcolor{blue}{0.403}} & 0.410 & 0.436 & 0.508 & 0.494 & 0.405 & 0.422 & 0.516 & 0.476 \\
\cmidrule(lr){2-18}
 & Avg. & 0.348 & \textbf{\textcolor{blue}{0.379}} & 0.347 & 0.389 & \textbf{\textcolor{blue}{0.346}} & 0.383 & \textbf{\textcolor{red}{0.343}} & \textbf{\textcolor{red}{0.369}} & 0.368 & 0.395 & 0.405 & 0.412 & 0.370 & 0.388 & 0.444 & 0.429 \\
\midrule
\multirow{5}{*}{\rotatebox[origin=c]{90}{ETTm1}} & 96 & \textbf{\textcolor{red}{0.290}} & \textbf{\textcolor{red}{0.324}} & 0.316 & 0.371 & \textbf{\textcolor{blue}{0.292}} & \textbf{\textcolor{blue}{0.326}} & 0.322 & 0.328 & 0.336 & 0.345 & 0.433 & 0.389 & 0.318 & 0.329 & 0.525 & 0.438 \\
 & 192 & \textbf{\textcolor{red}{0.338}} & \textbf{\textcolor{red}{0.356}} & 0.382 & 0.409 & \textbf{\textcolor{blue}{0.340}} & \textbf{\textcolor{blue}{0.358}} & 0.385 & 0.361 & 0.396 & 0.373 & 0.469 & 0.413 & 0.368 & \textbf{\textcolor{blue}{0.358}} & 0.616 & 0.484 \\
 & 336 & \textbf{\textcolor{red}{0.375}} & \textbf{\textcolor{red}{0.380}} & 0.447 & 0.441 & \textbf{\textcolor{blue}{0.377}} & \textbf{\textcolor{blue}{0.382}} & 0.429 & 0.385 & 0.437 & 0.400 & 0.517 & 0.439 & 0.395 & \textbf{\textcolor{red}{0.380}} & 0.678 & 0.516 \\
 & 720 & \textbf{\textcolor{red}{0.442}} & \textbf{\textcolor{blue}{0.418}} & 0.550 & 0.488 & \textbf{\textcolor{blue}{0.444}} & 0.420 & 0.500 & 0.423 & 0.510 & 0.434 & 0.532 & 0.610 & 0.455 & \textbf{\textcolor{red}{0.414}} & 0.756 & 0.553 \\
\cmidrule(lr){2-18}
 & Avg. & \textbf{\textcolor{red}{0.361}} & \textbf{\textcolor{red}{0.370}} & 0.424 & 0.427 & \textbf{\textcolor{blue}{0.363}} & \textbf{\textcolor{blue}{0.372}} & 0.409 & 0.374 & 0.420 & 0.388 & 0.488 & 0.463 & 0.384 & \textbf{\textcolor{red}{0.370}} & 0.644 & 0.498 \\
\midrule
\multirow{5}{*}{\rotatebox[origin=c]{90}{ETTm2}} & 96 & \textbf{\textcolor{red}{0.177}} & \textbf{\textcolor{blue}{0.246}} & 0.186 & 0.275 & \textbf{\textcolor{blue}{0.179}} & 0.256 & 0.189 & 0.248 & 0.208 & 0.272 & 0.209 & 0.277 & \textbf{\textcolor{blue}{0.179}} & \textbf{\textcolor{red}{0.244}} & 0.243 & 0.305 \\
 & 192 & \textbf{\textcolor{red}{0.241}} & \textbf{\textcolor{red}{0.293}} & 0.255 & 0.321 & \textbf{\textcolor{blue}{0.243}} & 0.300 & 0.261 & 0.295 & 0.274 & 0.318 & 0.297 & 0.329 & 0.252 & \textbf{\textcolor{blue}{0.294}} & 0.323 & 0.355 \\
 & 336 & \textbf{\textcolor{red}{0.294}} & \textbf{\textcolor{red}{0.331}} & 0.321 & 0.361 & \textbf{\textcolor{blue}{0.296}} & 0.336 & 0.317 & \textbf{\textcolor{blue}{0.333}} & 0.334 & 0.348 & 0.411 & 0.381 & 0.319 & 0.336 & 0.398 & 0.397 \\
 & 720 & \textbf{\textcolor{red}{0.374}} & \textbf{\textcolor{red}{0.384}} & 0.425 & 0.420 & \textbf{\textcolor{blue}{0.376}} & 0.388 & 0.395 & \textbf{\textcolor{blue}{0.386}} & 0.410 & 0.401 & 0.532 & 0.477 & 0.421 & 0.395 & 0.525 & 0.464 \\
\cmidrule(lr){2-18}
 & Avg. & \textbf{\textcolor{red}{0.272}} & \textbf{\textcolor{red}{0.314}} & 0.297 & 0.344 & \textbf{\textcolor{blue}{0.274}} & 0.320 & 0.291 & \textbf{\textcolor{blue}{0.316}} & 0.307 & 0.335 & 0.362 & 0.366 & 0.293 & 0.317 & 0.372 & 0.380 \\
\midrule
\multirow{5}{*}{\rotatebox[origin=c]{90}{ECL}} & 96 & 0.124 & \textbf{\textcolor{blue}{0.211}} & 0.140 & 0.240 & \textbf{\textcolor{red}{0.111}} & \textbf{\textcolor{red}{0.208}} & \textbf{\textcolor{blue}{0.122}} & \textbf{\textcolor{red}{0.208}} & $-$ & $-$ & 0.168 & 0.267 & $-$ & $-$ & 0.209 & 0.273 \\
 & 192 & 0.143 & 0.229 & 0.160 & 0.258 & \textbf{\textcolor{red}{0.127}} & \textbf{\textcolor{red}{0.225}} & \textbf{\textcolor{blue}{0.141}} & \textbf{\textcolor{blue}{0.226}} & $-$ & $-$ & 0.191 & 0.281 & $-$ & $-$ & 0.233 & 0.296 \\
 & 336 & 0.171 & 0.258 & 0.182 & 0.280 & \textbf{\textcolor{red}{0.145}} & \textbf{\textcolor{red}{0.244}} & \textbf{\textcolor{blue}{0.159}} & \textbf{\textcolor{blue}{0.245}} & $-$ & $-$ & 0.214 & 0.312 & $-$ & $-$ & 0.267 & 0.325 \\
 & 720 & \textbf{\textcolor{red}{0.195}} & \textbf{\textcolor{red}{0.280}} & 0.234 & 0.324 & \textbf{\textcolor{blue}{0.197}} & 0.289 & 0.202 & \textbf{\textcolor{blue}{0.282}} & $-$ & $-$ & 0.269 & 0.351 & $-$ & $-$ & 0.342 & 0.380 \\
\cmidrule(lr){2-18}
 & Avg. & 0.158 & 0.245 & 0.179 & 0.276 & \textbf{\textcolor{red}{0.145}} & \textbf{\textcolor{blue}{0.242}} & \textbf{\textcolor{blue}{0.156}} & \textbf{\textcolor{red}{0.240}} & $-$ & $-$ & 0.211 & 0.303 & $-$ & $-$ & 0.263 & 0.319 \\
\midrule
\multirow{5}{*}{\rotatebox[origin=c]{90}{Weather}} & 96 & \textbf{\textcolor{red}{0.155}} & \textbf{\textcolor{red}{0.190}} & 0.185 & 0.225 & \textbf{\textcolor{blue}{0.157}} & 0.204 & 0.161 & \textbf{\textcolor{blue}{0.192}} & 0.180 & 0.216 & 0.218 & 0.225 & $-$ & $-$ & 0.227 & 0.237 \\
 & 192 & \textbf{\textcolor{red}{0.207}} & \textbf{\textcolor{red}{0.236}} & 0.233 & 0.270 & \textbf{\textcolor{blue}{0.209}} & 0.252 & \textbf{\textcolor{blue}{0.209}} & \textbf{\textcolor{blue}{0.238}} & 0.230 & 0.268 & 0.290 & 0.274 & $-$ & $-$ & 0.290 & 0.290 \\
 & 336 & \textbf{\textcolor{red}{0.262}} & \textbf{\textcolor{red}{0.278}} & 0.299 & 0.316 & \textbf{\textcolor{blue}{0.264}} & 0.292 & 0.266 & \textbf{\textcolor{blue}{0.280}} & 0.279 & 0.308 & 0.356 & 0.328 & $-$ & $-$ & 0.350 & 0.334 \\
 & 720 & \textbf{\textcolor{red}{0.334}} & \textbf{\textcolor{red}{0.328}} & 0.384 & 0.370 & \textbf{\textcolor{blue}{0.336}} & 0.342 & 0.340 & \textbf{\textcolor{blue}{0.330}} & 0.349 & 0.357 & 0.547 & 0.454 & $-$ & $-$ & 0.429 & 0.388 \\
\cmidrule(lr){2-18}
 & Avg. & \textbf{\textcolor{red}{0.240}} & \textbf{\textcolor{red}{0.258}} & 0.275 & 0.295 & \textbf{\textcolor{blue}{0.242}} & 0.273 & 0.244 & \textbf{\textcolor{blue}{0.260}} & 0.260 & 0.287 & 0.353 & 0.320 & $-$ & $-$ & 0.324 & 0.312 \\
\midrule
\multirow{5}{*}{\rotatebox[origin=c]{90}{Solar}} & 96 & \textbf{\textcolor{blue}{0.163}} & \textbf{\textcolor{blue}{0.208}} & \textbf{\textcolor{red}{0.144}} & \textbf{\textcolor{red}{0.202}} & 0.165 & 0.210 & 0.300 & 0.252 & 0.309 & 0.271 & 0.686 & 0.673 & 0.213 & 0.209 & 0.686 & 0.690 \\
 & 192 & \textbf{\textcolor{blue}{0.187}} & \textbf{\textcolor{red}{0.227}} & \textbf{\textcolor{red}{0.182}} & 0.231 & 0.189 & \textbf{\textcolor{blue}{0.229}} & 0.341 & 0.275 & 0.360 & 0.285 & 0.699 & 0.697 & 0.245 & \textbf{\textcolor{blue}{0.229}} & 0.706 & 0.296 \\
 & 336 & \textbf{\textcolor{red}{0.208}} & 0.284 & 0.232 & 0.266 & \textbf{\textcolor{blue}{0.210}} & \textbf{\textcolor{red}{0.247}} & 0.346 & 0.282 & 0.357 & 0.292 & 0.728 & 0.710 & 0.281 & \textbf{\textcolor{blue}{0.251}} & 0.723 & 0.707 \\
 & 720 & \textbf{\textcolor{red}{0.245}} & 0.290 & 0.305 & 0.319 & \textbf{\textcolor{blue}{0.247}} & \textbf{\textcolor{blue}{0.274}} & 0.352 & 0.288 & 0.367 & 0.299 & 0.764 & 0.726 & 0.296 & \textbf{\textcolor{red}{0.263}} & 0.761 & 0.729 \\
\cmidrule(lr){2-18}
 & Avg. & \textbf{\textcolor{red}{0.201}} & 0.252 & 0.216 & 0.255 & \textbf{\textcolor{blue}{0.203}} & \textbf{\textcolor{blue}{0.240}} & 0.335 & 0.274 & 0.348 & 0.287 & 0.719 & 0.702 & 0.259 & \textbf{\textcolor{red}{0.238}} & 0.719 & 0.606 \\
\midrule
\multicolumn{2}{c|}{1st Count} & 20 & 16 & 4 & 1 & 3 & 4 & 1 & 4 & 0 & 0 & 0 & 0 & 0 & 5 & 0 & 0 \\
\bottomrule
\end{tabular}%
}
\endgroup
\end{table}

\section{Pretrained Time Series Forecasting}
\label{App_pretrained}

Table~\ref{tab:pretrained_full} presents the complete pretrained
forecasting results with a lookback length of 672 and prediction
horizons $\{96,192,336,720\}$. For each dataset, we report MSE and MAE at each prediction horizon and their averages over the four horizons. Pre-training and evaluation details are provided in Appendix~\ref{App_implementation}. Datasets included in a model's pre-training data are excluded from its evaluation and marked with a dash. UniTFU achieves the lowest average MSE on five datasets and the lowest average MAE on four. In particular, it achieves the lowest MSE at all four prediction horizons on ETTh1, ETTm1, ETTm2, and Weather. Across the 28 forecasting settings, UniTFU obtains the lowest MSE in 20 settings and the lowest MAE in 16. Performance nevertheless varies across datasets and metrics: TiRex achieves lower average MSE and MAE on ETTh2 and ECL, while TimesFM and Sundial achieve lower average MAE on Solar.

\begin{figure}[h]
    \centering
    \includegraphics[width=\linewidth]{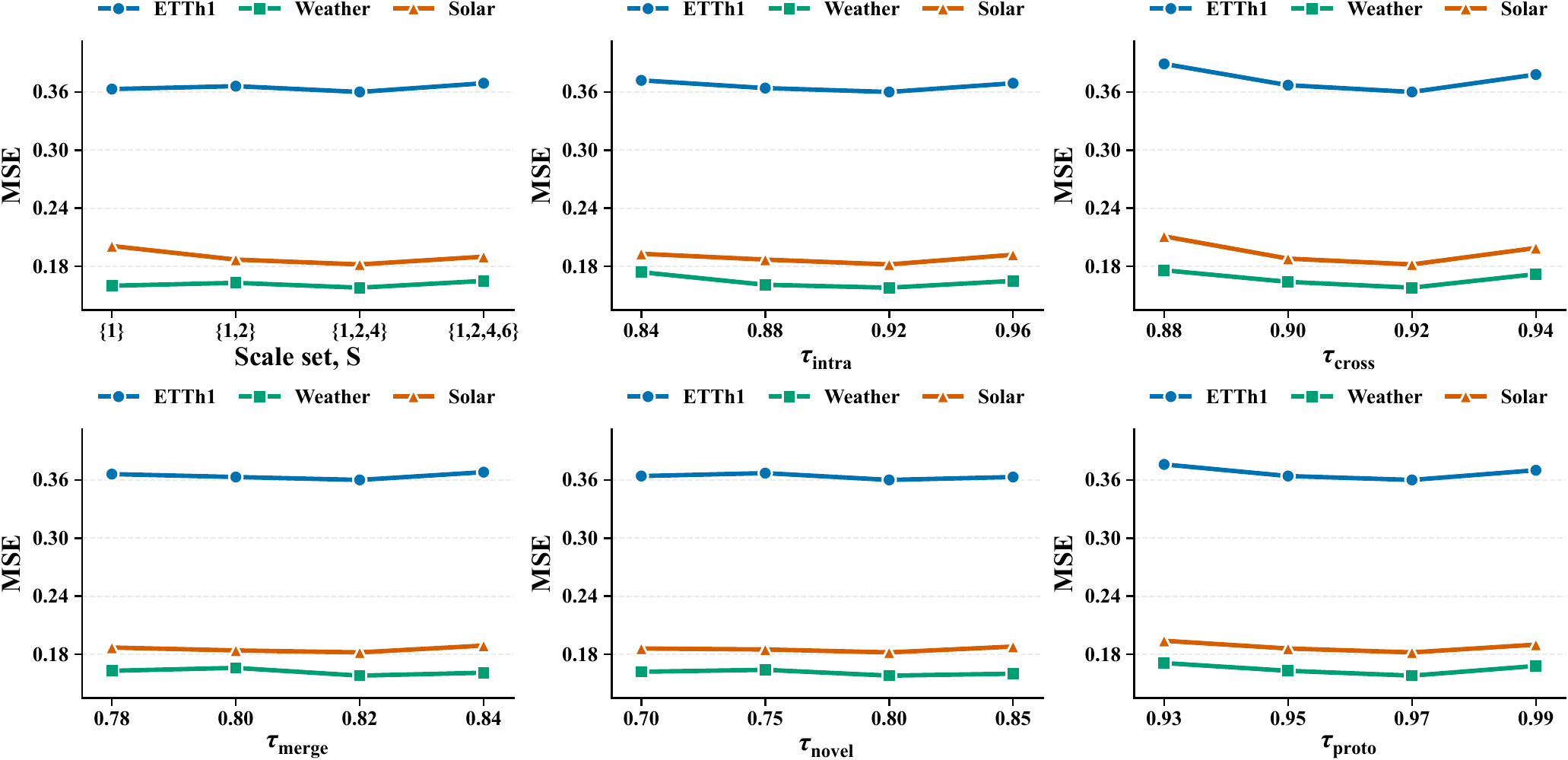}
    \caption{
    Hyperparameter sensitivity of UniTFU on ETTh1, Weather, and Solar under the $672\rightarrow96$ forecasting setting.
    The top row varies the temporal scale set $\mathcal{S}$, the intra-scale redundancy threshold $\tau_{\mathrm{intra}}$, and the cross-scale redundancy threshold $\tau_{\mathrm{cross}}$.
    The bottom row varies the candidate routing threshold $\tau_{\mathrm{merge}}$, the novelty grouping threshold $\tau_{\mathrm{novel}}$, and the prototype consolidation threshold $\tau_{\mathrm{proto}}$.
    We maintain $\tau_{\mathrm{new}}=\tau_{\mathrm{merge}}-0.10$ and fix all other hyperparameters.
    }
    \label{fig:hyperparameter_sensitivity}
\end{figure}

\section{Hyperparameter Sensitivity}
\label{App_sensitivity}

We evaluate the sensitivity of UniTFU to the main hyperparameters involved in TFU discovery and dictionary maintenance on ETTh1, Weather, and Solar under the $672\rightarrow96$ forecasting setting. As shown in Figure~\ref{fig:hyperparameter_sensitivity}, we vary six factors individually: the temporal scale set $\mathcal{S}$, the intra-scale and cross-scale redundancy thresholds $\tau_{\mathrm{intra}}$ and $\tau_{\mathrm{cross}}$, the candidate routing threshold $\tau_{\mathrm{merge}}$, the novelty grouping threshold $\tau_{\mathrm{novel}}$, and the prototype consolidation threshold $\tau_{\mathrm{proto}}$. For each factor, all other hyperparameters are fixed to the reference configuration described in Appendix~\ref{App_implementation}. For candidate routing, we maintain the fixed margin $\tau_{\mathrm{new}}=\tau_{\mathrm{merge}}-0.10$ rather than tuning the two thresholds independently. All configurations are trained using the same random seed, and MSE is reported. Across the three datasets, the reference configuration consistently attains the lowest MSE among the tested settings. For the temporal scale set, progressively extending $\mathcal{S}$ from $\{1\}$ to $\{1,2,4\}$ improves forecasting accuracy, whereas further introducing scale 6 leads to a moderate degradation. This indicates that combining local dynamics at multiple temporal scales is beneficial, while excessively long candidate segments do not provide additional gains under the current forecasting context. The threshold analyses show a similar pattern. For redundancy suppression, both $\tau_{\mathrm{intra}}$ and $\tau_{\mathrm{cross}}$ achieve their lowest errors at $0.92$, while either more permissive or more restrictive filtering degrades performance. The effect is particularly visible for $\tau_{\mathrm{cross}}$, where lowering the threshold to $0.88$ consistently increases the error across all three datasets. For TFU dictionary maintenance, the reference values $\tau_{\mathrm{merge}}=0.82$, $\tau_{\mathrm{novel}}=0.80$, and $\tau_{\mathrm{proto}}=0.97$ also yield the lowest MSE within their respective tested ranges. Moderate threshold perturbations yield relatively small MSE
changes within the tested ranges. Overall, the same reference configuration performs consistently across datasets with different temporal characteristics, supporting our use of a shared set of TFU-specific hyperparameters rather than dataset-specific tuning.

\begin{table}[t]
\centering
\caption{Complete ablation results under the $672\rightarrow96$
forecasting setting. Each row reports results for one dataset
and random seed.}
\label{tab:ablation_full}
\begingroup
\setlength{\tabcolsep}{3pt}
\renewcommand{\arraystretch}{1.08}
\resizebox{\textwidth}{!}{%
\begin{tabular}{ll|rr|rr|rr|rr|rr|rr}
\toprule
\multirow{2}{*}{Dataset} & \multirow{2}{*}{Seed}
& \multicolumn{2}{c|}{UniTFU}
& \multicolumn{2}{c|}{w/o TFU}
& \multicolumn{2}{c|}{w/o Sce}
& \multicolumn{2}{c|}{w/o Att}
& \multicolumn{2}{c|}{w/o Res}
& \multicolumn{2}{c}{w/ VQ} \\
& & MSE & MAE & MSE & MAE & MSE & MAE
& MSE & MAE & MSE & MAE & MSE & MAE \\
\midrule
\multirow{4}{*}{ETTh1} & 2021 & \textbf{0.360} & \textbf{0.396} & 0.366 & 0.404 & 0.363 & 0.399 & 0.365 & 0.400 & 0.367 & 0.399 & 0.364 & 0.401 \\
 & 2023 & 0.362 & 0.399 & 0.369 & 0.409 & \textbf{0.360} & \textbf{0.395} & 0.371 & 0.405 & 0.372 & 0.403 & 0.364 & 0.406 \\
 & 2025 & \textbf{0.361} & \textbf{0.399} & 0.363 & 0.401 & 0.366 & 0.402 & 0.368 & 0.405 & 0.371 & 0.403 & 0.365 & 0.401 \\
 & 2027 & 0.364 & 0.402 & 0.370 & 0.405 & \textbf{0.361} & 0.400 & 0.373 & \textbf{0.399} & 0.375 & 0.406 & 0.367 & 0.402 \\
\midrule
\multirow{4}{*}{ETTh2} & 2021 & \textbf{0.297} & \textbf{0.350} & 0.299 & 0.353 & 0.300 & 0.356 & 0.303 & 0.358 & 0.301 & 0.354 & 0.298 & 0.352 \\
 & 2023 & \textbf{0.294} & \textbf{0.346} & 0.303 & 0.358 & 0.304 & 0.360 & 0.297 & 0.350 & 0.306 & 0.360 & 0.301 & 0.355 \\
 & 2025 & \textbf{0.299} & \textbf{0.351} & 0.304 & 0.357 & 0.304 & 0.360 & 0.307 & 0.363 & 0.304 & 0.357 & 0.302 & 0.355 \\
 & 2027 & 0.302 & \textbf{0.354} & 0.301 & 0.355 & \textbf{0.299} & \textbf{0.354} & 0.311 & 0.366 & 0.310 & 0.362 & 0.304 & 0.360 \\
\midrule
\multirow{4}{*}{ETTm1} & 2021 & \textbf{0.272} & \textbf{0.327} & 0.287 & 0.335 & 0.279 & 0.331 & 0.274 & 0.328 & 0.273 & 0.329 & 0.276 & 0.330 \\
 & 2023 & 0.275 & 0.331 & 0.291 & 0.340 & 0.282 & 0.332 & \textbf{0.270} & \textbf{0.325} & 0.280 & 0.337 & 0.283 & 0.337 \\
 & 2025 & \textbf{0.270} & \textbf{0.326} & 0.290 & 0.339 & 0.283 & 0.334 & 0.271 & \textbf{0.326} & 0.279 & 0.336 & 0.282 & 0.333 \\
 & 2027 & \textbf{0.276} & 0.332 & 0.288 & 0.337 & 0.278 & \textbf{0.329} & 0.284 & 0.336 & 0.282 & 0.339 & 0.285 & 0.336 \\
\midrule
\multirow{4}{*}{ETTm2} & 2021 & \textbf{0.182} & \textbf{0.265} & 0.188 & 0.269 & 0.186 & 0.268 & 0.183 & \textbf{0.265} & 0.184 & 0.266 & 0.187 & 0.267 \\
 & 2023 & 0.185 & 0.269 & 0.186 & 0.268 & 0.190 & 0.273 & \textbf{0.181} & \textbf{0.264} & 0.188 & 0.271 & 0.188 & 0.269 \\
 & 2025 & \textbf{0.180} & \textbf{0.264} & 0.191 & 0.273 & 0.190 & 0.271 & 0.186 & 0.269 & 0.182 & 0.265 & 0.188 & 0.270 \\
 & 2027 & \textbf{0.185} & 0.269 & 0.189 & 0.271 & \textbf{0.185} & \textbf{0.266} & 0.186 & 0.269 & 0.186 & 0.269 & 0.186 & 0.272 \\
\midrule
\multirow{4}{*}{ECL} & 2021 & \textbf{0.132} & \textbf{0.229} & 0.135 & 0.233 & 0.133 & 0.231 & 0.136 & 0.234 & 0.135 & 0.232 & 0.134 & 0.230 \\
 & 2023 & \textbf{0.128} & \textbf{0.226} & 0.139 & 0.238 & 0.136 & 0.235 & 0.140 & 0.239 & 0.133 & 0.229 & 0.136 & 0.233 \\
 & 2025 & 0.134 & 0.230 & 0.139 & 0.236 & \textbf{0.130} & \textbf{0.229} & 0.141 & 0.238 & 0.138 & 0.236 & 0.135 & 0.234 \\
 & 2027 & 0.136 & 0.238 & 0.138 & 0.235 & 0.135 & 0.234 & \textbf{0.134} & \textbf{0.231} & 0.139 & 0.237 & 0.137 & 0.234 \\
\midrule
\multirow{4}{*}{Weather} & 2021 & 0.158 & \textbf{0.206} & 0.162 & 0.213 & 0.159 & 0.208 & \textbf{0.156} & 0.207 & 0.157 & 0.208 & 0.161 & 0.211 \\
 & 2023 & 0.162 & 0.211 & 0.166 & 0.218 & 0.162 & 0.212 & \textbf{0.154} & \textbf{0.206} & 0.160 & 0.212 & 0.161 & 0.213 \\
 & 2025 & 0.160 & 0.209 & 0.158 & 0.210 & 0.164 & 0.214 & 0.159 & 0.211 & \textbf{0.154} & \textbf{0.206} & 0.162 & 0.212 \\
 & 2027 & \textbf{0.155} & \textbf{0.205} & 0.163 & 0.215 & 0.158 & 0.206 & 0.160 & 0.210 & 0.160 & 0.212 & 0.161 & 0.211 \\
\midrule
\multirow{4}{*}{Solar} & 2021 & \textbf{0.182} & \textbf{0.220} & 0.191 & 0.226 & 0.195 & 0.230 & 0.197 & 0.231 & 0.188 & 0.225 & 0.183 & 0.222 \\
 & 2023 & \textbf{0.179} & \textbf{0.218} & 0.194 & 0.230 & 0.198 & 0.232 & 0.193 & 0.228 & 0.192 & 0.228 & 0.190 & 0.226 \\
 & 2025 & \textbf{0.184} & 0.223 & 0.193 & 0.227 & 0.192 & 0.228 & 0.201 & 0.236 & 0.186 & \textbf{0.222} & \textbf{0.184} & 0.223 \\
 & 2027 & \textbf{0.184} & 0.223 & 0.189 & 0.225 & 0.196 & 0.233 & 0.200 & 0.235 & 0.186 & \textbf{0.222} & 0.190 & 0.225 \\
\bottomrule
\end{tabular}%
}
\endgroup
\end{table}

\section{Full Ablation Study Results}
\label{App_ablation}

Table~\ref{tab:ablation_full} reports the individual ablation results on all seven datasets under the $672\rightarrow96$ forecasting setting, using seeds $\{2021,2023,2025,2027\}$. Seed 2021 is used for the benchmark experiments. Each row reports the MSE and MAE from one run. The variants are defined in Section~\ref{exp_ablat}.

\noindent\textbf{VQ Codebook Replacement.}
The w/ VQ variant replaces the adaptive TFU dictionary with a fixed-size codebook while retaining the same multi-scale candidate encoding, redundancy suppression, token-level TFU composition, scene context aggregation, TFU attention, forecasting head, and historical-pattern residual. The codebook contains $K=128$ codewords with the same embedding dimension as the TFU prototypes, matching the maximum capacity of the adaptive dictionary rather than its number of active TFUs. Codewords are initialized from the first non-empty batch of retained training candidates. When at least $K$ candidates are available, $K$ candidates are sampled without replacement; otherwise, all available candidates are retained and additional candidates are sampled with replacement to fill the codebook.

Each retained candidate is assigned to the codeword with the highest cosine similarity, without confidence-based routing. During training, assigned candidate embeddings are stored in an update buffer that retains at most the 4096 most recent assignments. At each codebook maintenance step, embeddings assigned to the same codeword are averaged, and the corresponding codeword is updated by grouped EMA using the same momentum $\eta=0.95$ as the adaptive dictionary. Codewords without buffered assignments remain unchanged. Candidate codeword matching follows the same normalization setting as the adaptive dictionary, and the resulting codeword representations are used by the same token-level composition procedure. Unlike the adaptive TFU dictionary, the VQ codebook has a fixed size and does not perform confidence-based routing, novelty birth, or prototype consolidation. Codewords are maintained as non-gradient buffers rather than learnable parameters, and no additional VQ commitment or reconstruction loss is introduced.

\section{Additional Model Analysis}
\label{App_model_analysis}

\begin{figure*}[t]
    \centering
    \includegraphics[width=0.90\linewidth]
    {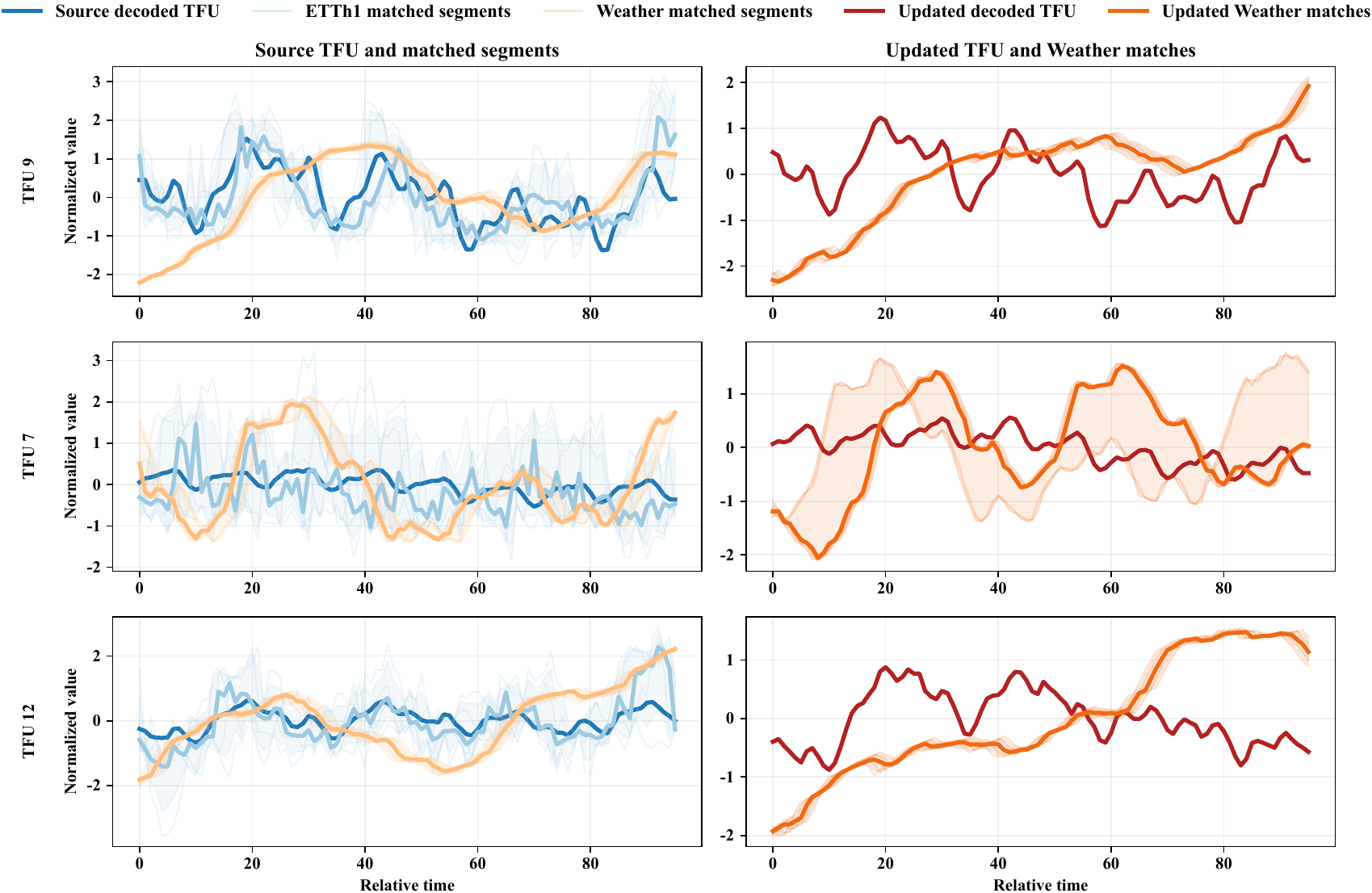}
    \caption{
    Representative TFUs and matched temporal segments from
    ETTh1 to Weather.
    Left: decoded source-domain TFUs and matched segments
    from both domains.
    Right: decoded updated TFUs and matched Weather segments.
    }
    \label{fig:tfu_semantics}
\end{figure*}

\noindent\textbf{Representative TFU Semantics.}
We visualize representative TFUs and their matched temporal segments in the ETTh1$\rightarrow$Weather setting. A lightweight post-hoc decoder maps latent prototypes to one-dimensional temporal curves. As shown in Fig.~\ref{fig:tfu_semantics}, the left column compares source-domain TFUs with matched ETTh1 and Weather segments, while the right column compares updated TFUs with their Weather matches. These examples provide a qualitative view of the local dynamics associated with individual TFUs, rather than exact reconstructions of the matched segments. The decoder is used only for visualization and does not participate in forecasting or TFU learning.

\noindent\textbf{Shape Analysis Settings.}
We compare nearest-neighbor retrieval using learned embeddings or raw temporal statistics, together with random retrieval. The raw-statistics baseline computes the mean, maximum, and endpoint difference directly from each raw segment, whereas the semantic encoder summarizes learned token representations. Self-matches and overlapping segments are excluded. Waveform correlation is computed after linear detrending and normalization to control for level, scale, and linear trend. For each retrieval method, we first average correlations over anchors within each TFU and then compute the mean and standard deviation across TFUs. This gives each TFU equal weight rather than allowing frequently activated TFUs to dominate the reported results. The bars in Fig.~\ref{fig:embedding_shape} report these across-TFU means, and the error bars indicate across-TFU
standard deviations. The paired bootstrap 95\% confidence intervals assess the mean correlation improvement of embedding-based retrieval over raw-statistics retrieval; they are distinct from the plotted error bars. The Weather setting uses the source-trained ETTh1 TFU dictionary. Updated Weather uses the dictionary updated and expanded with limited Weather training data, while retaining the same Weather evaluation data. The number of active TFUs increases from 16 to 22. Dictionary updating changes prototype assignments and TFU groups, thereby changing the weighting of anchors in the across-TFU average. Consequently, the mean correlation of raw-statistics retrieval changes slightly from 0.656 to 0.664, although its definition remains unchanged. For the extended cross-domain evaluation, we additionally include frequency-based nearest-neighbor retrieval. All retrieval strategies follow the same candidate eligibility constraints and waveform correlation evaluation. Frozen source dictionaries are compared with their versions updated using limited target-domain training data.

\begin{figure}[h]
    \centering
    \includegraphics[width=0.72\linewidth]
    {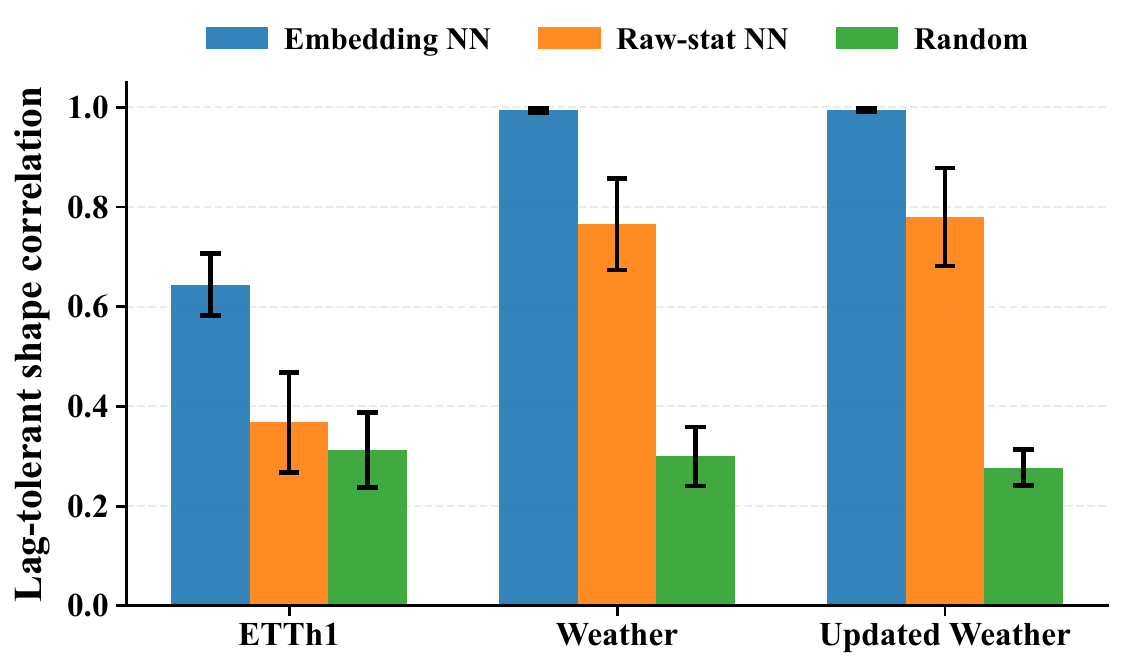}
    \caption{
    Waveform correlation after linear detrending and normalization,
    allowing temporal shifts of up to $\pm8$ steps.
    Bars report the mean of within-TFU average correlations;
    error bars indicate the standard deviation across TFUs.
    }
    \label{fig:embedding_shape_lag}
\end{figure}

\noindent\textbf{Additional Shape Analysis.}
We further evaluate waveform correlation while allowing temporal shifts of up to $\pm8$ steps. As shown in Fig.~\ref{fig:embedding_shape_lag}, embedding-based retrieval achieves the highest mean correlations: 0.645, 0.994, and 0.995 on ETTh1, Weather, and updated Weather, respectively, compared with 0.368, 0.766, and 0.780 for raw-statistics retrieval. The advantage persists when small temporal shifts are allowed, further supporting the local shape information retained in the learned embeddings.

\noindent\textbf{Within-TFU Shape Consistency.}
We examine whether segments assigned to the same TFU exhibit similar local shapes. Table~\ref{tab:tfu_shape_consistency} compares waveform correlations between same-TFU segments and randomly paired segments after linear detrending and normalization. Correlations are averaged within each TFU and then summarized across TFUs. Same TFU segments have higher mean correlations in all three settings, although consistency varies across TFUs and is weaker on ETTh1. On Weather, the mean correlation increases from 0.372 to 0.456 after dictionary updating, while random-pair correlations remain close to zero. Under the corresponding TFU assignments, these results indicate greater average within-TFU shape consistency after target-domain adaptation.

These results support local shape retention under the evaluated conditions, rather than complete preservation of temporal ordering or invariance to arbitrary temporal distortions.

\begin{table}[t]
    \centering
    \caption{
    Within-TFU shape consistency after linear detrending
    and normalization.
    Values report the mean $\pm$ standard deviation
    of within-TFU average correlations across TFUs.
    }
    \label{tab:tfu_shape_consistency}
    \setlength{\tabcolsep}{7pt}
    \renewcommand{\arraystretch}{1.05}
    \begin{tabular}{lcc}
        \toprule
        Setting & Same TFU & Random \\
        \midrule
        ETTh1
        & $0.069 \pm 0.078$
        & $0.007 \pm 0.009$ \\
        Weather
        & $0.372 \pm 0.355$
        & $0.014 \pm 0.064$ \\
        Updated Weather
        & $0.456 \pm 0.359$
        & $0.019 \pm 0.047$ \\
        \bottomrule
    \end{tabular}
\end{table}

\begin{table*}[h]
\centering
\caption{Cross-domain local shape correlations using frozen and updated TFU dictionaries. All correlations are computed after linear detrending and normalization. Higher values indicate greater waveform similarity.}
\label{tab:cross_domain_shape_retrieval}
\setlength{\tabcolsep}{7pt}
\renewcommand{\arraystretch}{1.05}
\begin{tabular}{l|l|c|c|c|c}
\toprule
Source $\rightarrow$ Target
& Dictionary
& Embedding NN
& Frequency NN
& Raw-stat NN
& Random \\
\midrule
\multirow{2}{*}{ETTh1 $\rightarrow$ Weather}
& Frozen
& \textbf{0.789} & 0.679 & 0.656 & 0.040 \\
& Updated
& \textbf{0.825} & 0.694 & 0.664 & 0.041 \\
\midrule
\multirow{2}{*}{ETTh1 $\rightarrow$ ETTm1}
& Frozen
& \textbf{0.688} & 0.606 & 0.564 & 0.023 \\
& Updated
& \textbf{0.726} & 0.609 & 0.565 & 0.028 \\
\midrule
\multirow{2}{*}{ECL $\rightarrow$ Solar}
& Frozen
& \textbf{0.699} & 0.618 & 0.568 & 0.042 \\
& Updated
& \textbf{0.747} & 0.633 & 0.598 & 0.046 \\
\midrule
\multirow{2}{*}{Weather $\rightarrow$ ECL}
& Frozen
& 0.585 & \textbf{0.607} & 0.533 & 0.035 \\
& Updated
& \textbf{0.604} & 0.592 & 0.542 & 0.040 \\
\bottomrule
\end{tabular}
\end{table*}

\noindent\textbf{Cross-domain Shape Retrieval.}
We extend the local shape analysis to four source--target pairs using frozen and updated TFU dictionaries. As shown in Table~\ref{tab:cross_domain_shape_retrieval}, embedding-based retrieval achieves the highest detrended waveform correlation in 7 of 8 settings, outperforming frequency-based, raw-statistics, and random retrieval in most cases.

\noindent\textbf{Prototype Usage and Predictive Relevance.}
We examine the relationship between prototype usage and window-level forecasting errors across all 20 source--target transfers under the $672\rightarrow96$ setting. Table~\ref{tab:prototype_predictive_relevance} reports the within-transfer Spearman correlations, changes in mean best-match cosine similarity, and forecasting errors before and after dictionary updating.

\begin{table*}[t]
\centering
\caption{Prototype usage, matching similarity changes, and forecasting performance across 20 source--target adaptations. Usage--Error $\rho$ denotes the within-transfer Spearman correlation between window-level prototype usage and forecasting MSE. $\Delta$Similarity measures the change in mean matching cosine similarity, and $\Delta$MSE denotes Frozen MSE minus Updated MSE. Positive $\Delta$MSE indicates forecasting improvement. The Overall row reports the median Usage--Error correlation and means for the remaining columns.}
\label{tab:prototype_predictive_relevance}
\small
\setlength{\tabcolsep}{7pt}
\renewcommand{\arraystretch}{1.05}
\begin{tabular}{l|c|c|c|c|c}
\toprule
Source $\rightarrow$ Target
& Usage--Error $\rho$
& $\Delta$Similarity
& Frozen MSE
& Updated MSE
& $\Delta$MSE \\
\midrule

ETTm1$\rightarrow$ETTh1
& $-0.131$ & $+0.012$ & 0.447 & 0.445 & $+0.002$ \\
Weather$\rightarrow$ETTh1
& $-0.087$ & $+0.018$ & 0.566 & 0.557 & $+0.009$ \\
ECL$\rightarrow$ETTh1
& $+0.033$ & $+0.015$ & 0.355 & 0.362 & $-0.007$ \\
Solar$\rightarrow$ETTh1
& $-0.203$ & $+0.028$ & 0.559 & 0.546 & $+0.013$ \\

\midrule

ETTh1$\rightarrow$ETTm1
& $-0.144$ & $+0.016$ & 0.426 & 0.421 & $+0.005$ \\
Weather$\rightarrow$ETTm1
& $-0.093$ & $+0.018$ & 0.446 & 0.433 & $+0.013$ \\
ECL$\rightarrow$ETTm1
& $+0.047$ & $+0.013$ & 0.436 & 0.440 & $-0.004$ \\
Solar$\rightarrow$ETTm1
& $-0.213$ & $+0.024$ & 0.478 & 0.464 & $+0.014$ \\

\midrule

ETTh1$\rightarrow$Weather
& $-0.216$ & $+0.022$ & 0.188 & 0.184 & $+0.004$ \\
ETTm1$\rightarrow$Weather
& $-0.123$ & $+0.015$ & 0.184 & 0.181 & $+0.003$ \\
ECL$\rightarrow$Weather
& $-0.074$ & $+0.008$ & 0.192 & 0.185 & $+0.007$ \\
Solar$\rightarrow$Weather
& $-0.184$ & $+0.009$ & 0.197 & 0.194 & $+0.003$ \\

\midrule

ETTh1$\rightarrow$ECL
& $-0.147$ & $+0.014$ & 0.217 & 0.211 & $+0.006$ \\
ETTm1$\rightarrow$ECL
& $-0.257$ & $+0.028$ & 0.232 & 0.215 & $+0.017$ \\
Weather$\rightarrow$ECL
& $+0.023$ & $+0.010$ & 0.233 & 0.217 & $+0.016$ \\
Solar$\rightarrow$ECL
& $-0.223$ & $+0.035$ & 0.256 & 0.223 & $+0.033$ \\

\midrule

ETTh1$\rightarrow$Solar
& $-0.285$ & $+0.030$ & 0.266 & 0.240 & $+0.026$ \\
ETTm1$\rightarrow$Solar
& $-0.266$ & $+0.026$ & 0.250 & 0.228 & $+0.022$ \\
Weather$\rightarrow$Solar
& $-0.176$ & $+0.018$ & 0.241 & 0.223 & $+0.018$ \\
ECL$\rightarrow$Solar
& $-0.247$ & $+0.033$ & 0.280 & 0.240 & $+0.040$ \\

\midrule
\textbf{Overall}
& $\mathbf{-0.162}$
& $\mathbf{+0.020}$
& \textbf{0.322}
& \textbf{0.310}
& $\mathbf{+0.012}$ \\

\bottomrule
\end{tabular}
\end{table*}

Prototype usage is negatively correlated with forecasting MSE in 17 of 20 transfers, with a median Spearman correlation of $-0.162$. Mean matching similarity increases in all 20 transfers, while dictionary updating reduces MSE in 18. Across transfers, similarity gains are positively associated with MSE reductions (Spearman $\rho=0.721$). However, ECL$\rightarrow$ETTh1 and ECL$\rightarrow$ETTm1 show increased similarity but worse forecasting errors. These results indicate that better prototype matching is associated with predictive improvements across transfers, but does not guarantee lower prediction error in every setting.

\end{document}